\documentclass[runningheads]{llncs}
\usepackage{graphicx}

\usepackage{tikz}
\usepackage{comment}
\usepackage{amsmath,amssymb} 
\usepackage{color}

\usepackage[colorlinks,linkcolor=red,anchorcolor=blue,citecolor=green]{hyperref}
\usepackage{float}
\usepackage{subfigure}
\usepackage{multirow}
\usepackage{marvosym}
\usepackage{ifsym}

\usepackage[accsupp]{axessibility}  

\usepackage[width=122mm,left=12mm,paperwidth=146mm,height=193mm,top=12mm,paperheight=217mm]{geometry}

\begin{document}
\pagestyle{headings}
\mainmatter
\def\ECCVSubNumber{5430}  

\title{EleGANt: Exquisite and Locally Editable GAN for Makeup Transfer} 

\titlerunning{EleGANt: Exquisite and Locally Editable GAN for Makeup Transfer}
%
\author{Chenyu Yang\inst{1}\and
Wanrong He\inst{1}\and
Yingqing Xu\inst{1}\textsuperscript{(\textrm{\Letter})}\and
Yang Gao\inst{1,2}\textsuperscript{(\textrm{\Letter})}
}

%
\authorrunning{C. Yang et al.}
%
\institute{Tsinghua University, Beijing, China\\
\email{\{yangcy19, hwr19\}@mails.tsinghua.edu.cn\\ 
\{yqxu, gaoyangiiis\}@tsinghua.edu.cn} \and
Shanghai Qi Zhi Institute, Shanghai, China
}
\maketitle
\let\thefootnote\relax\footnotetext{\textsuperscript{\textrm{\Letter}}Corresponding author}

\begin{figure}[!htbp]
\vskip -0.2cm
\setlength{\abovecaptionskip}{-0.1cm}
\setlength{\belowcaptionskip}{-0.6cm} 
\centering
\includegraphics[width=0.96\textwidth]{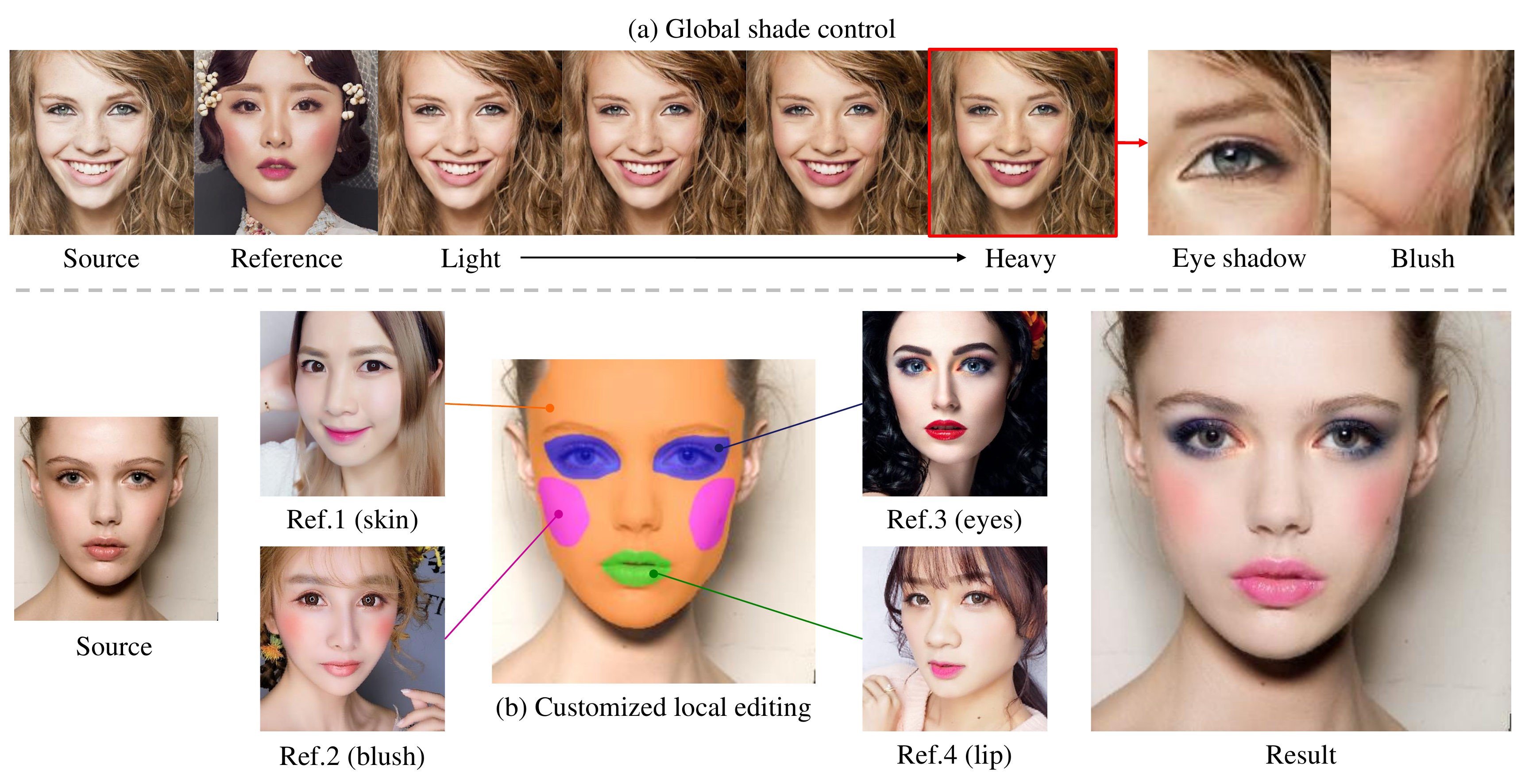}
\caption{Our proposed EleGANt generates makeup faces with exquisite details. It supports flexible control such as (a) global shade control and (b) customized local editing.}
\label{fig-teaser}
\end{figure}

\begin{abstract}
Most existing methods view makeup transfer as transferring color distributions of different facial regions and ignore details such as eye shadows and blushes. Besides, they only achieve controllable transfer within predefined fixed regions. This paper emphasizes the transfer of makeup details and steps towards more flexible controls. To this end, we propose Exquisite and locally editable GAN for makeup transfer (EleGANt). 
It encodes facial attributes into pyramidal feature maps to preserves high-frequency information. It uses attention to extract makeup features from the reference and adapt them to the source face, and we introduce a novel Sow-Attention Module that applies attention within shifted overlapped windows to reduce the computational cost. Moreover, EleGANt is the first to achieve customized local editing within arbitrary areas by corresponding editing on the feature maps. Extensive experiments demonstrate that EleGANt generates realistic makeup faces with exquisite details and achieves state-of-the-art performance. The code is available at \href{https://github.com/Chenyu-Yang-2000/EleGANt}{https://github.com/Chenyu-Yang-2000/EleGANt}. 
\keywords{Makeup Transfer, Sow-Attention, GAN Controls}
\end{abstract}

\section{Introduction}
Makeup transfer aims to transfer the makeup from an specific reference image to a source image. It has tremendous value in practical scenarios, for instance, cosmetics try-on and marketing. The goal of makeup transfer is two-fold: (1) Precisely transferring the makeup attributes from the reference to the source. The attributes include low-frequency color features and high-frequency details such as the brushes of eye shadow and blushes on the cheek. (2) Preserving the identity of the source, involving shapes, illuminations, and even subtle wrinkles. Controllable transfer further meets the requirement in practice: it allows users to design customized makeup according to their preferences.

Deep learning approaches, especially GAN-based models \cite{BeautyGAN,PairedCycleGAN,LADN,PSGAN,SCGAN}, have been widely employed in this task. They mainly adopt the CycleGAN \cite{CycleGAN} framework that is trained on unpaired non-makeup and with-makeup images, with extra supervision, e.g., a makeup loss term \cite{BeautyGAN,PairedCycleGAN,FAT}, to guide the reconstruction of a specific makeup on the source face. 
Notwithstanding the demonstrated success, existing approaches mainly view makeups as color distributions and largely ignore the spatial, high-frequency information about details. Some methods \cite{BeautyGAN,LADN} cannot tackle the pose misalignment between the two faces, while others represent makeups by matrices of limited size \cite{PSGAN} or 1D-vectors \cite{SCGAN}, resulting in high-frequency attributes being smoothed.
Meanwhile, the commonly used objective, histogram matching \cite{BeautyGAN}, only imposes constraints on color distributions without incorporating any spatial information. Besides, existing controllable models \cite{PSGAN,SCGAN} only achieve editing the makeup within some fixed regions, for instance, skin, lip, and eyes, instead of arbitrary customized regions.

To address these issues, we propose Exquisite and locally editable GAN for makeup transfer (EleGANt). On one hand, we focus on high-frequency information to synthesize makeup faces with rich and delicate details.
EleGANt encodes facial attributes of different frequencies into feature maps of a pyramid structure. To tackle misaligned head poses, it uses QKV-attention to extract makeup features from the reference and adapt them to the source face by pixel-wise correspondence. 
We employ a high-resolution feature map to preserve high-frequency attributes and further propose a novel Sow-Attention Module to reduce the computational cost. It computes attention efficiently within local windows and uses shifted overlapped windowing schemes to ensure the continuity of the output. The network is trained with a newly designed pseudo ground truth which comprises both color and spatial information.
On the other hand, the high-res makeup feature maps support precise editing to control the makeup style and shade within arbitrary customized areas. To our best knowledge, EleGANt is the first makeup transfer network to achieve this free-style local editing.

Fig. \ref{fig-teaser} exhibits the great capability and controllability of EleGANt. It generates realistic makeup images with high-fidelity colors and high-quality details. Global shade control (Fig. \ref{fig-teaser}(a)) and customized local editing (Fig. \ref{fig-teaser}(b)) can be realized by manipulating the makeup feature maps. Extensive experiments demonstrate the superiority of EleGANt compared with the existing methods.

The contributions of this paper can be summarized as follows:
\begin{itemize}
    \item We propose EleGANt, a fully automatic makeup transfer network with the most flexible control among existing methods. To our best knowledge, it is the first to achieve customized local editing of makeup style and shade.
    \item EleGANt uses a pyramid structure with a high-resolution feature map to preserve high-frequency makeup features beyond color distributions. It achieves state-of-the-art performance, especially in processing makeup details.
    \item A novel Sow-Attention Module that computes attention within shifted overlapped windows is introduced, which guarantees the continuity of the output and reduces the computational cost for high-resolution inputs.
\end{itemize}

\section{Related Work}
\subsection{Makeup Transfer}
Makeup transfer has been studied in computer vision for a decade. Traditional methods \cite{Example-1,Example-2,tradition-1,tradition-2,tradition-3} utilized image processing techniques. Later, CycleGAN \cite{CycleGAN} and its variants \cite{StarGAN,FDIT} were widely used for image-to-image translation tasks such as facial attribute transfer. However, these methods focus on domain-level rather than instance-level transfer, and they do not well maintain some facial attributes, e.g., shape and pose, that keep unchanged during makeup.

Inspired by the successes in GANs, makeup transfer was formulated as an asymmetric domain translation problem in PairedCycleGAN \cite{PairedCycleGAN}. They also employed an additional discriminator to guide makeup transfer with pseudo transferred images. 
BeautyGAN \cite{BeautyGAN} introduced a dual input/output GAN for simultaneous makeup transfer and removal and a color histogram matching loss for instance-level makeup transfer. BeautyGlow \cite{BeautyGlow} utilized the Glow framework to disentangle the latent features into makeup and non-makeup components. LADN \cite{LADN} leveraged multiple and overlapping local discriminators to ensure local details consistency. PSGAN \cite{PSGAN} and FAT \cite{FAT} proposed to use attention mechanism to handle misaligned facial poses and expressions. Lately, SCGAN \cite{SCGAN} attempted to eliminate the spatial misalignment problem by encoding makeup styles into component-wise style-codes. 

However, the existing methods have limitations in processing makeup details: some methods \cite{BeautyGAN,LADN} cannot tackle the pose misalignment; \cite{SCGAN} encodes the makeups into 1D-vectors, thus discarding a large proportion of spatial information; the high cost of pixel-wise attention in \cite{PSGAN,FAT} limited the size of feature maps then harmed the preservation of details.
EleGANt surpasses these approaches by using a high-res feature map and an efficient Sow-Attention Module. Besides, unlike previous models \cite{PSGAN,SCGAN} that can only adjust the makeup in a fixed set of regions, our EleGANt supports customized local editing in arbitrary regions.

\subsection{Style Transfer}
Style transfer can be regarded as a general form of makeup transfer, and it has been investigated extensively since the rise of deep convolutional neural networks \cite{DIA,NeuralTransfer,PerceptualLoss,phototransfer}. However, these methods either require a time-consuming optimization process or can only transfer a fixed set of styles. 
Then \cite{AdaIN} proposed adaptive instance normalization (AdaIN) that matched the mean and variance of the content features with those of the style features and achieved arbitrary style transfer. Since style transfer methods do not consider the face-specific semantic correspondence and lack local manipulation and controllability, even the state-of-the-art algorithms \cite{style-1,style-2,style-3} cannot fit makeup transfer applications.

\subsection{Attention Mechanism}

Attention mechanism has been widely used in the computer vision area. Early works \cite{vattn-1,vattn-2} simply apply it to images where each pixel attends to every other pixel, but they do not scale to large input sizes due to a quadratic cost in the number of pixels. Many variants have been tried so far to make attention more efficient and applicable to high-resolution images, e.g., tokenization on patches \cite{ViT,T2T}, attention in local regions \cite{local-attn-1,Swin,ViL,Twins}, and pyramid architecture \cite{PVT,PiT,R2L}. Since these modules are plugged into discriminative networks for image representations, they are not exactly suitable for generative tasks. Meanwhile, sparse attention such as block-wise \cite{block-1,block-2}, axial-wise \cite{axial-1,GODIVA}, and nearby attention \cite{NUWA} has been introduced into visual synthesis models, but neither global nor regional attributes of the makeup can be completely encoded due to the sparsity.

For makeup transfer, naive pixel-wise attention employed by \cite{PSGAN,FAT} suffers from significant computational overhead and is hard to scale to inputs of larger size. To address this problem, EleGANt uses a novel Sow-Attention Module that performs attention within shifted overlapped windows.

\section{Methodology}
\subsection{Formulation}
Let $X$ and $Y$ be the non-makeup image domain and the
makeup image domain, where $\{x^n\}_{n=1,\dots,N},x^n\in X$ and $\{y^m\}_{m=1,\dots,M},y^m\in Y$ denote the examples of two domains respectively. We assume no paired data is available, i.e., the non-makeup and makeup images have different identities. Our proposed EleGANt aims to learn a transfer function $\mathcal{G}$: given a source image $x$ and a reference image $y$, $\hat{x}=\mathcal{G}(x,y)$, where $\hat{x}$ is the transferred image with the makeup style of $y$ and the face identity of $x$.

\subsection{Network Architecture}
\label{sec:arch}
\begin{figure}[t]
    \setlength{\belowcaptionskip}{-0.2cm} 
    \centering
    \includegraphics[width=0.75\textwidth]{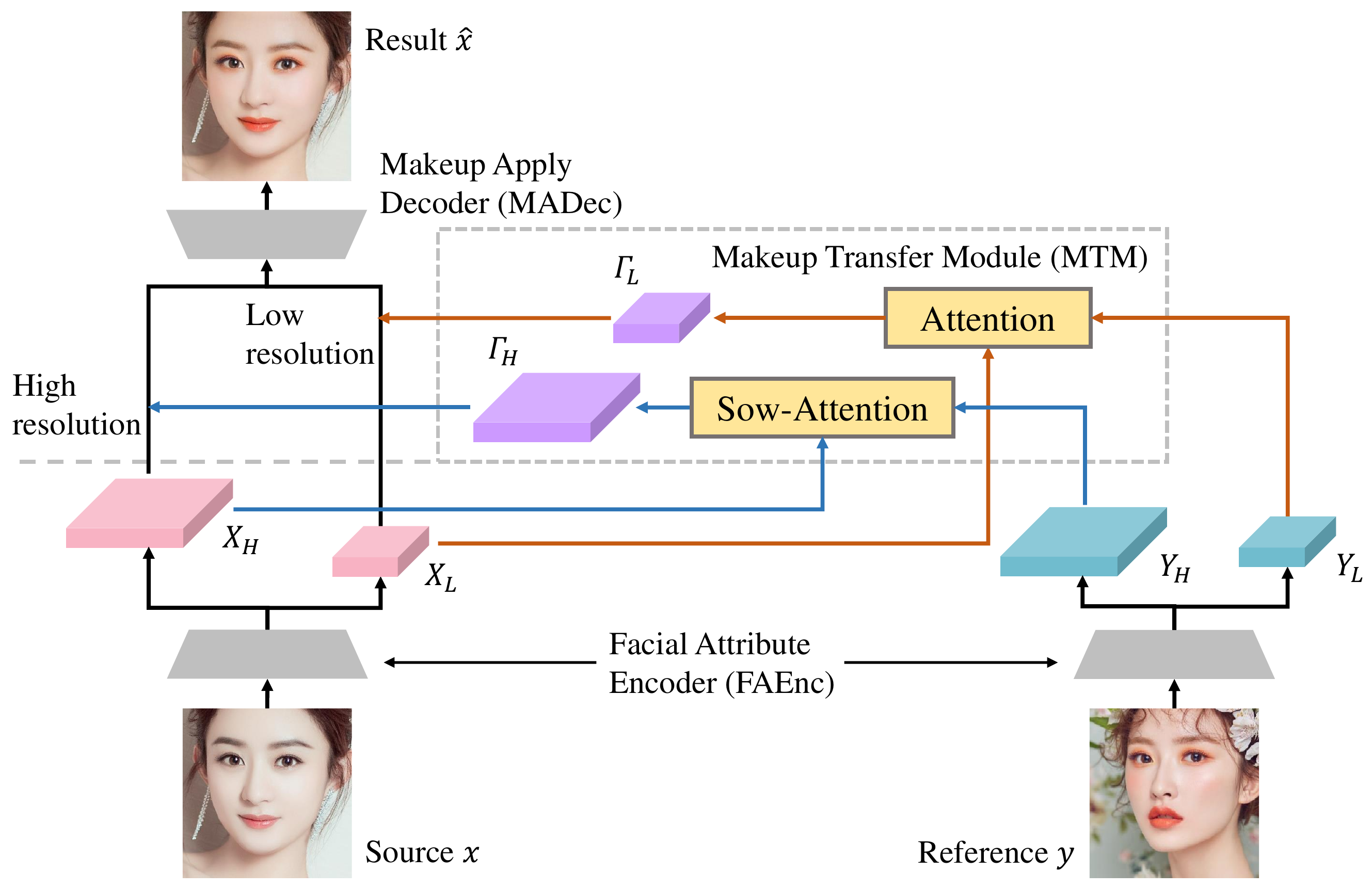}
\caption{Overall structure of our proposed EleGANt. Facial Attribute Encoder (FAEnc) constructs pyramidal feature maps. Makeup Transfer Module (MTM) yields low-res and high-res makeup feature maps, $\Gamma_L$ and $\Gamma_H$, by Attention and Sow-Attention modules respectively. Makeup Apply Decoder (MADec) applies the makeup feature maps to the source to generate the final result.}
    \label{fig-arch}
\end{figure}
\subsubsection{Overall.}
The architecture of EleGANt is shown in Fig. \ref{fig-arch}, which consists of three components: (1) \textit{Facial Attribute Encoder.} FAEnc encodes the facial attributes into feature maps of pyramid structure. High-res feature maps $X_H$, $Y_H$ mainly contain high-frequency information about sharp edges and details, while low-res ones $X_L$, $Y_L$ contain more low-frequency information related to colors and shadows. (2) \textit{Makeup Transfer Module}. In MTM, an Attention Module is applied to the low-res feature maps and yields a low-res makeup feature map $\Gamma_L$, while a Sow-Attention Module is for the high-res one $\Gamma_H$. These two modules extract makeup features of different frequencies and utilize attention to make them spatially aligned with the source face to tackle the misalignment between the two faces. (3) \textit{Makeup Apply Decoder.} MADec applies the two makeup feature maps $\Gamma_L$ and $\Gamma_H$ to corresponding feature maps of the source respectively by element-wise multiplication and generates the final result. 

\subsubsection{Attention Module.} 
Since the two faces may have discrepancies in expressions and poses, the makeup attributes of the reference face need to be adapted to the source face. We employ a QKV-cross-attention similar to Transformer \cite{Transformer} to model the pixel-wise correspondence between the two faces. Formally, given a pair of feature maps extracted from the source and reference, $X,Y\in \mathbb{R}^{HW\times C}$, where $C$, $H$ and $W$ are the number of channels, height and width of the feature map, we compute the attentive matrix $A\in \mathbb{R}^{HW\times HW}$ to specify how a pixel on $X$ corresponds to its counterparts on $Y$:
\begin{equation}
\begin{aligned}
    A=softmax\left(\frac{\widetilde{X}Q(\widetilde{Y}K)^T}{\sqrt{C}}\right)
\end{aligned}
\end{equation}
where $K,Q\in\mathbb{R}^{C\times C}$ 
are learnable parameters, 
and $\widetilde{X},\widetilde{Y}$ are the feature maps combined with positional embedding to introduce spatial information. Here, we adopt Landmark Embedding \cite{FAT} that concatenates a vector representing the relative positions to the facial landmarks with the visual features for each pixel.

Makeup features are extracted by a $1\times1$-Conv with weights $V\in\mathbb{R}^{C\times C}$ from $Y$. They consequently maintain the spatial correspondence with $Y$. Then the attentive matrix $A$ is applied to align them with the spatial distribution of $X$. The aligned features are represented by a makeup feature map $\Gamma\in\mathbb{R}^{HW\times C}$:
\begin{equation}
\begin{aligned}
    \Gamma =A(YV)
\end{aligned}
\end{equation}
After that, the makeup feature map $\Gamma$ becomes the input of MADec and then is applied to the source feature map $X$ by element-wise multiplication: 
\begin{equation}
\begin{aligned}
    \widehat{X} = \Gamma \odot X
\end{aligned}   
\end{equation}
Before being fed into MADec, $\Gamma$ can be globally or locally manipulated to achieve controllable transfer, which will be discussed in detail in Sec. \ref{sec:control}.

\subsubsection{Sow-Attention Module.} 
\begin{figure}[t]
    \setlength{\belowcaptionskip}{-0.1cm}
    \centering
    \includegraphics[width=0.99\textwidth]{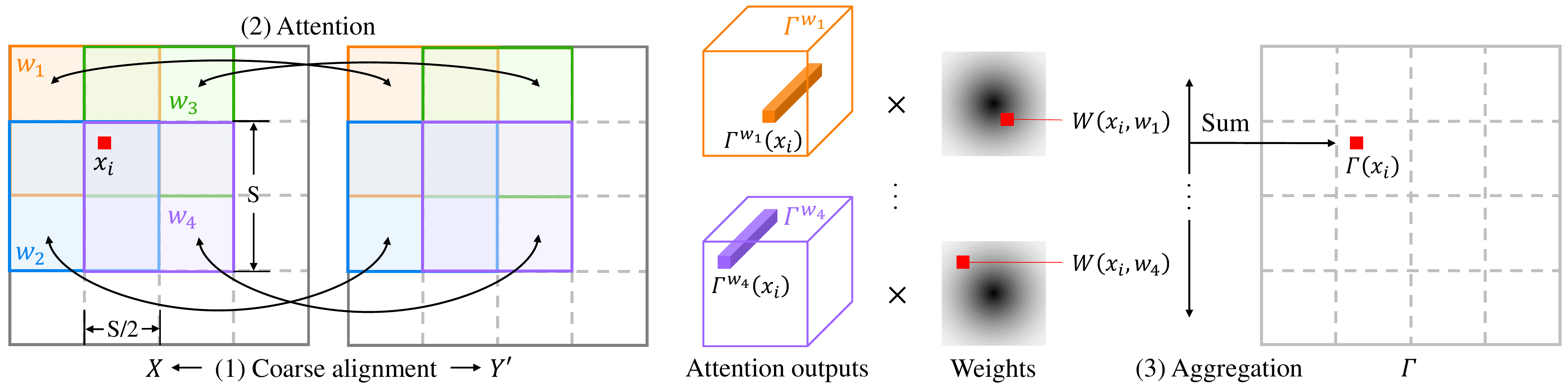}
\caption{Illustration of the Sow-Attention Module. Attention is computed within shifted overlapped windows across coarsely aligned feature maps, and the outputs are aggregated by weighted sum. A darker color indicates a larger weight.}
\label{fig:sow}
\end{figure}

To avoid high-frequency information being smoothed and support precise local editing, we utilize feature maps of high resolution. However, the above pixel-wise attention is not practicable here due to the high quadratic cost. We propose a novel shifted overlapped windowing attention (Sow-Attention) to reduce the complexity. As illustrated in Fig. \ref{fig:sow}, the Sow-Attention Module obtains the makeup feature map $\Gamma$ by three steps:

(1) \textit{Coarse alignment.} We employ Thin Plate Splines (TPS) to warp  $Y$ into $Y'$ to be coarsely aligned with the source face. Specifically, TPS is determined by $N$ control points whose coordinates in the original space and the target space are denoted as $C = \{c_i\}_{i=1}^N$ and $C' = \{c'_i\}_{i=1}^N$, respectively. Here, $C$ is set to be the coordinates of $N$ landmark points \cite{landmark} of the reference face and $C'$ to be those of the source face, and then TPS warps $Y$ into $Y'$ to fit $C$ into $C'$. We use the parameterized grid sampling \cite{STN} to perform 2D TPS that is differentiable w.r.t input $Y$ and the formulation from \cite{ASTER} to obtain required parameters.

(2) \textit{Attention.} Since $X$ and $Y'$ are coarsely aligned, local attention is enough for a point on $X$ to capture makeup information from the neighbor region on $Y'$. To avoid the boundary issue of non-overlapped windows that leads to artificial edges in the output image (see Sec. \ref{sec:ablation}), we perform attention in shifted overlapped windows. As depicted in Fig. \ref{fig:sow}, $w_1$, $w_2$, $w_3$ and $w_4$ represent 4 partitioning schemes that split the feature maps the into overlapped windows of size $S$ with an $S/2$ shift. A QKV-attention is shared by all windows, and the result computed within window $w_j$ is denoted as $\Gamma^{w_j}\in\mathbb{R}^{S^2\times C}$, where $C$ is the number of channels. An alternative view is that a pixel cross-attends to the four windows it belongs to, for instance, the pixel $x_i$ marked in Fig. \ref{fig:sow} attends to the windows $\{w_j\}_{j=1}^4$ on $Y'$ and obtains four vectors $\Gamma^{w_j}(x_i)\in \mathbb{R}^{C},j=1,2,3,4$.

(3) \textit{Aggregation.} For each pixel $x_i$, the four vectors derived from previous attention, $\{\Gamma^{w_j}(x_i),\ j:x_i\in w_j\}$, are aggregated into one vector as the final output. We conduct this by a weighted sum:
\begin{equation}
\begin{aligned}
\Gamma(x_i)=\sum_{j:x_i\in w_j}\Gamma^{w_j}(x_i)\cdot W(x_i,w_j)
\end{aligned}
\label{eq:sum}
\end{equation}
where the weight $W(x_i,w_j)$ is determined by the relative position of $x_i$ to $w_j$. $W(x_i,w_j)$ should guarantee that the output is spatially continuous both inter and intra windows. Besides, if $x_i$ is closer to the center of $w_j$, $\Gamma^{w_j}(x_i)$ will contain more information about its neighbor region, and $W(x_i,w_j)$ is expected to be larger. We choose a ``bilinear'' form that works well in practice:
\begin{equation}
\begin{aligned}
    W(x_i,w_j)=\frac{\left|\left(S-2\left(\mathrm{x}(x_i)-\mathrm{x}(c_{w_j})\right)\right)\left(S-2\left(\mathrm{y}(x_i)-\mathrm{y}(c_{w_j})\right)\right)\right|}{S^2}
\end{aligned}
\label{eq:weight}
\end{equation}
where $c_{w_j}$ denotes the center of window $w_j$, $\mathrm{x}(\cdot)$ and $\mathrm{y}(\cdot)$ is the x-coordinate and y-coordinate of a point respectively. Eq. \ref{eq:sum} and Eq. \ref{eq:weight} can also be interpreted as a kind of ``bilinear interpolation'': the attention results from different windows are ``interpolated'' regarding their centers as anchor points. 


Sow-Attention reduces the cost of pixel-wise attention from $O\left((HW)^2\right)$ to $O\left(HWS^2\right)$. Generally, $S=H/8$ would be enough, and then the complexity is reduced by a factor of 16 since the attention is performed with four partitioning schemes and the cost of each is $(1/8)^2$ of the original attention.

\subsection{Makeup Loss with Pseudo Ground Truth}
\begin{figure}[t]
    \setlength{\belowcaptionskip}{-0.2cm}
    \centering
    \includegraphics[width=0.79\textwidth]{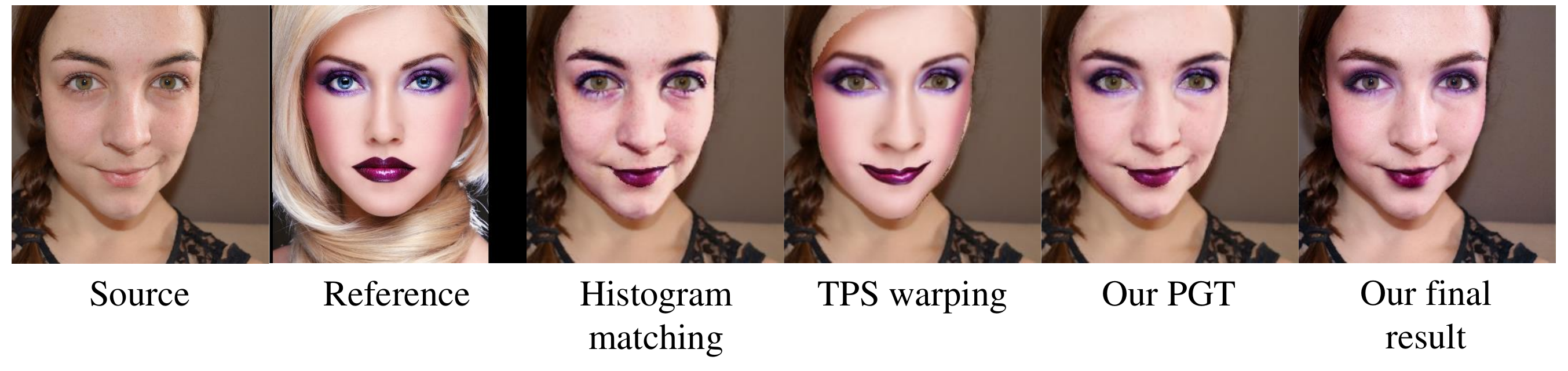}
    \caption{Different strategies for pseudo ground truth (PGT) generation. Our PGT has more accurate details (e.g., eye shadows) than histogram matching and fewer artifacts and distortions than TPS warping. The final generated image has better quality than PGTs, e.g., there are no jags on the edge of the lip and artifacts on the forehead.}
    \label{fig:PGT_Diff_strat}
\end{figure}

Due to the lack of paired makeup and non-makeup images, we adopt the CycleGAN \cite{CycleGAN} framework to train the network in an unsupervised way. Nevertheless, normal GAN training only drives the generator to produce realistic images with a general makeup. Therefore, to guide the reconstruction of specific makeup attributes of the reference on the source face, extra supervision, i.e., a makeup loss term for the generator, is introduced into the total objective $L_{total}$:
\begin{equation}
\begin{aligned}
    L_{total}&=\lambda_{a d v} (L_{\mathcal{G}}^{a d v} +  L_{\mathcal{D}}^{a d v})+\lambda_{c y c} L_{\mathcal{G}}^{c y c}+\lambda_{per} L_{\mathcal{G}}^{per}+\lambda_{make} L_{\mathcal{G}}^{make}
\end{aligned}
\end{equation}
where $L^{a d v},L^{c y c},L^{per}$ are adversarial loss \cite{GAN}, cycle consistency loss \cite{CycleGAN}, and perceptual loss \cite{PerceptualLoss} (formulations are summarized in App. 
\ref{sec:objective}). 
The makeup loss $L_{\mathcal{G}}^{make}$ is defined with pseudo ground truth (PGT), namely, images that are synthesized independently of the generator and serve as the training objective:
\begin{equation}
\begin{aligned}
L_{\mathcal{G}}^{m a k e} &=\|\mathcal{G}(x, y)-P G T(x, y)\|_{1} +\|\mathcal{G}(y, x)-P G T(y, x)\|_{1} .
\end{aligned}
\end{equation}
where $PGT(x,y)$ has the makeup of $y$ and the face identity of $x$. 
As shown in Fig. \ref{fig:PGT_Diff_strat}, typical strategies for PGT generation include histogram matching \cite{BeautyGAN,PSGAN} and TPS warping \cite{FAT}. Histogram matching equalizes the color distribution of the source face with that of the reference, but it suffers from extreme color differences and discards all spatial information. TPS warps the reference face into the shape of the source face by aligning detected landmarks, but it may result in artifacts of stitching and distortion and also mix in unwanted shadows. The imprecision of these PGTs will consequently cause a sub-optimal transfer. 

\begin{figure}[!htbp]
    \setlength{\belowcaptionskip}{-0.2cm}
    \centering
    \includegraphics[width=0.92\textwidth]{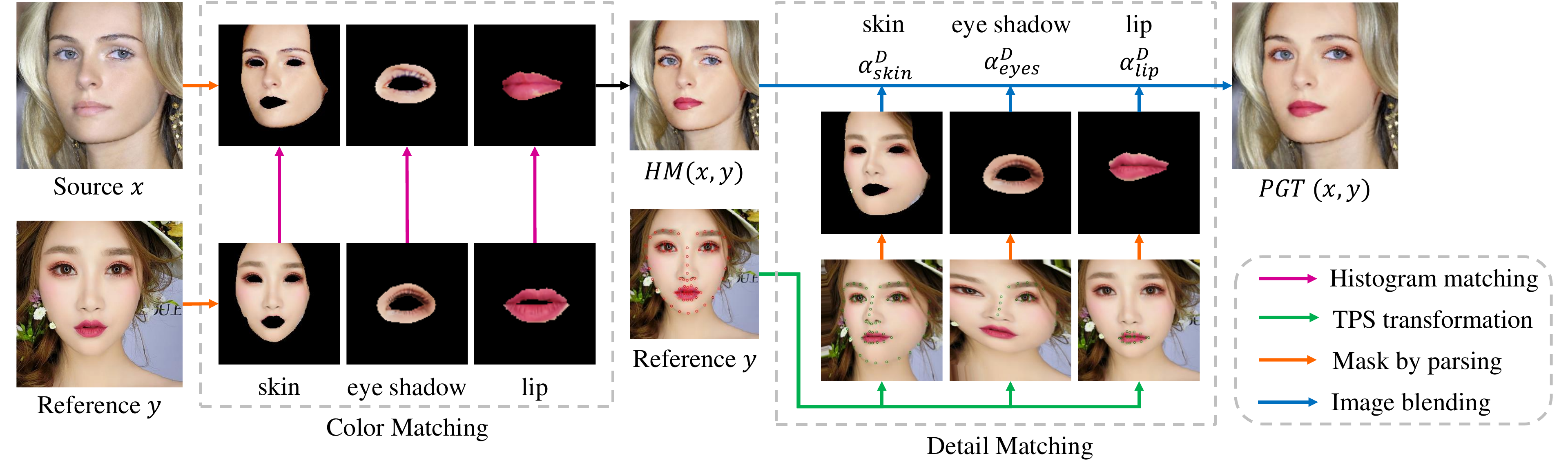}
    \caption{The pipeline of our PGT generation. It utilizes histogram matching for color matching, TPS warping for detail matching, and annealing factors for image blending.}
    \label{fig:PGT}
\end{figure}
To address these issues, we propose a novel strategy that incorporates both color and spatial information and avoids misleading signals. Although our PGT is not comparable with generated images in quality, it provides sufficient guidance complementary to the GAN training. As illustrated in Fig. \ref{fig:PGT}, it consists of two stages: color matching and detail matching.

(1) \textit{Color Matching.}
We adopt histogram matching \cite{BeautyGAN} to replicate the makeup color of the reference $y$ to the source $x$. The color distributions in the skin, lip, and eye shadow regions are separately equalized between $x$ and $y$.

(2) \textit{Detail Matching.}
We employ TPS transformation to incorporate spatial information into the PGT. Specifically, skin, lip, and eye shadow regions of the reference $y$ are separately warped to fit the source $x$ using corresponding facial landmarks, and blended with their counterparts on the color matching result. The blending factors $\alpha^D_{skin}$, $\alpha^D_{eyes}$, and $\alpha^D_{lip}$ anneal during the training process to emphasize colors or details in different stages of training for better results. See App. 
\ref{sec:pgt} 
for implementation details.

\section{Experiments}
\subsection{Experiment Settings}
We use the MT (Makeup Transfer) dataset \cite{BeautyGAN} which contains 1115 non-makeup images and 2719 makeup images to train our model. We follow the strategy of \cite{BeautyGAN} to split the train/test set. All images are resized to $256\times256$ before training. More implementation details and results are given in supplementary materials.

We conduct comparisons of EleGANt with general style transfer methods: DIA \cite{DIA}, CycleGAN \cite{CycleGAN} as well as makeup transfer methods: BeautyGAN \cite{BeautyGAN}, BeautyGlow \cite{BeautyGlow}, LADN \cite{LADN}, PSGAN \cite{PSGAN}, Spatial FAT \cite{FAT}, and SCGAN \cite{SCGAN}. Since the implementation of BeautyGlow and Spatial FAT is not released, we follow \cite{BeautyGAN} and take the results from corresponding papers. Table \ref{tab:func} summarizes the functions of open-source makeup transfer models. Our EleGANt demonstrates the greatest capability and flexibility among all methods. It can precisely transfer makeup details and is the first to achieve customized local editing.

\begin{table}[!htbp]
\setlength{\belowdisplayskip}{0cm}
\setlength{\abovecaptionskip}{0cm}
\centering
\caption{Analyses of EleGANt with existing open-source methods. ``Misalign.'': robust transfer with large spatial misalignment between the two faces. ``Detail'': precise transfer with high-quality details. ``Shade'': shade-controllable transfer. ``Part'': partial transfer for lip, eye, and skin regions. ``Local.'': local editing within arbitrary areas.}
\label{tab:func}
\begin{tabular}{c|cc|ccc}
\hline
\multirow{2}{*}{Method} & \multicolumn{2}{c|}{Capability}        & \multicolumn{3}{c}{Controllability}                            \\ \cline{2-6} 
                        & \multicolumn{1}{c|}{\;Misalign.\;} & \;Detail\; & \multicolumn{1}{c|}{\;Shade\;} & \multicolumn{1}{c|}{\;Part\;} & \;Local.\; \\ \hline
BeautyGAN \cite{BeautyGAN}\;   &  &  &  &  &  \\
LADN \cite{LADN}\;   &  &  & \checkmark &  &  \\
PSGAN \cite{PSGAN}\; & \checkmark &  & \checkmark & \checkmark &  \\
SCGAN \cite{SCGAN}\;  & \checkmark &  & \checkmark & \checkmark &  \\
EleGANt (ours)\; & \checkmark & \checkmark & \checkmark & \checkmark & \checkmark \\ \hline
\end{tabular}
\vskip -0.7cm
\end{table}

\subsection{Qualitative Comparison.}
\begin{figure}[t]
    \setlength{\abovedisplayskip}{0cm}
    \setlength{\abovecaptionskip}{-0.2cm}
    \setlength{\belowcaptionskip}{0.1cm}
    \centering
    \includegraphics[width=1.0\textwidth]{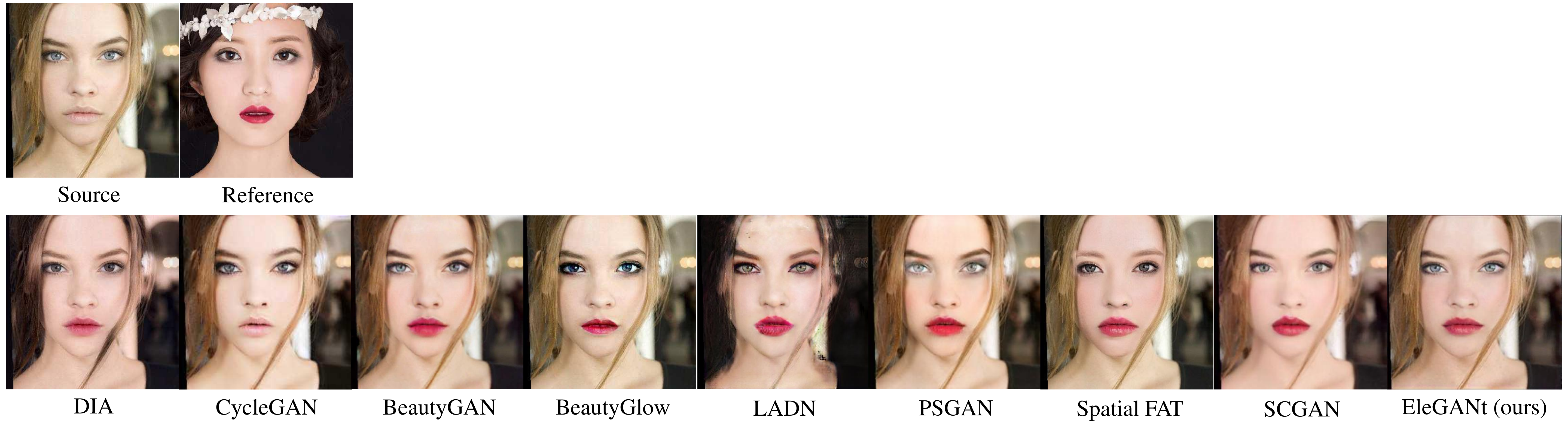}
    \caption{Qualitative comparisons with existing methods. EleGANt generates the most precise transferred result with the desired makeup and high-quality details.}
    \label{fig:model_comparison}
\end{figure}

Fig. \ref{fig:model_comparison} shows the results on images with frontal faces in neutral expressions and light makeups. The results of DIA have unnatural colors on the hair and shadows on the face. CycleGAN synthesizes realistic images, but it simply performs domain transfer without recovering any makeup of the reference. BeautyGlow fails to transfer the correct color of skin and lip. Severe artifacts and blurs exist in the results of LADN. Recent works such as PSGAN, SCGAN, and Spatial FAT can generate more visually acceptable results. However, PSGAN and SCGAN suffer from the color bleeding problem, e.g., blurs at the edge of the lips. Spatial FAT generates results with richer details, but there are still apparent artifacts. Compared to existing methods, our proposed EleGANt generates the most realistic images with natural light and shadow while synthesizing high-fidelity makeup colors and high-quality details.

\begin{figure}[!t]
    \setlength{\abovedisplayskip}{0cm}
    \setlength{\belowcaptionskip}{-0.2cm}
    \centering
    \includegraphics[width=0.90\textwidth]{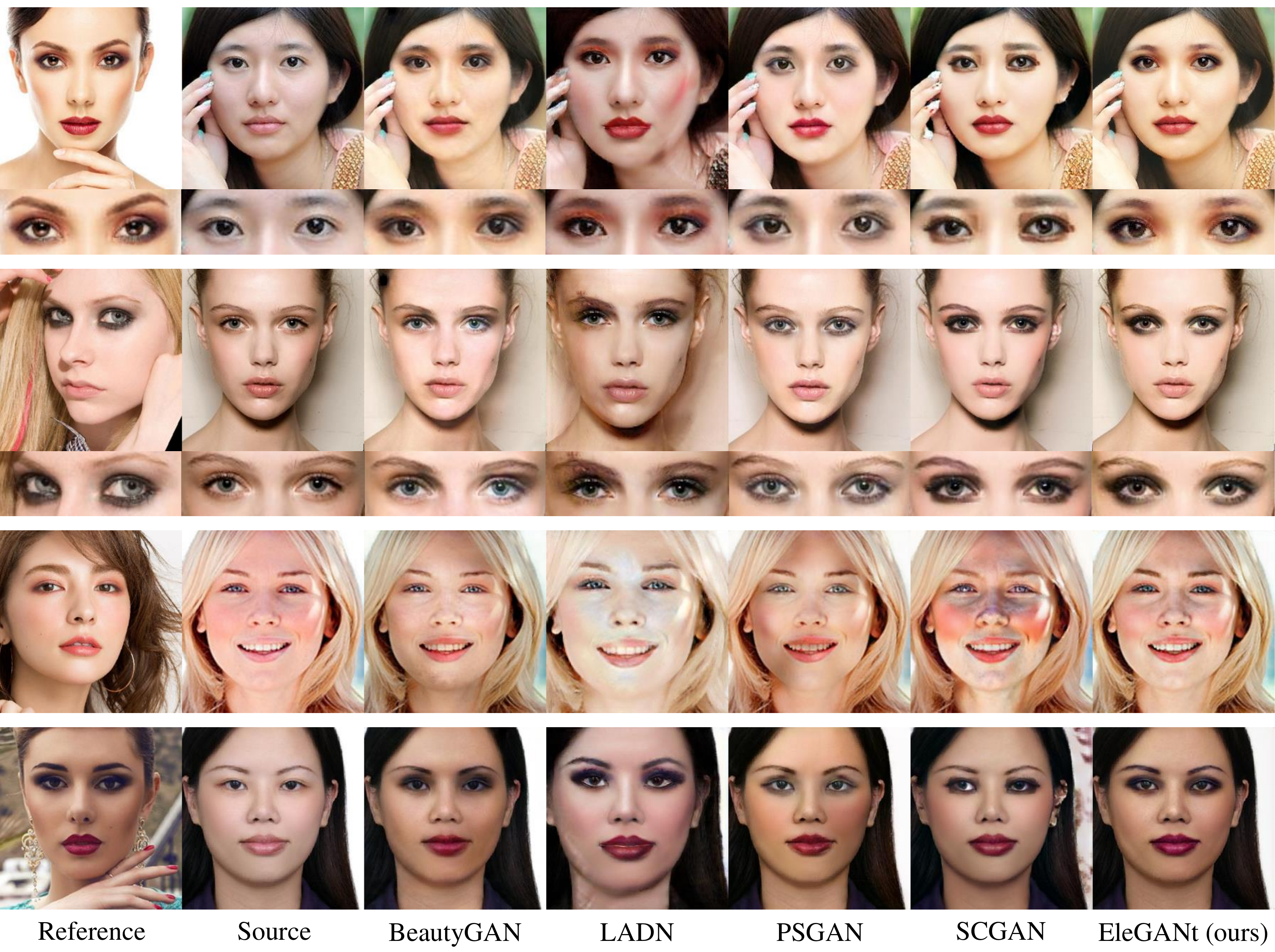}
    \caption{Comparison on images with misaligned poses and complex makeups. Our proposed EleGANt generates the most exquisite details (e.g., the shapes and colors of the eye shadows), the most natural colors and shadows, and the fewest artifacts.}
    \label{fig:robuts_comparison}
\end{figure}

To test the effectiveness on complex makeups and robustness against spatial misalignment, we compare our method with makeup transfer models that have released code. The results are shown in Fig. \ref{fig:robuts_comparison}. None of the existing methods can precisely transfer makeup details such as the shapes and colors of eye shadows (the 1st, 2nd, and 4th rows) due to the loss of high-frequency information. Our EleGANt synthesizes these details with the highest quality with the help of high-resolution feature maps. Existing methods fall short in cases that the two faces have large discrepancies in pose or illumination, resulting in unnatural colors and shadows (the 3rd and 4th rows) in the transferred images. The results of EleGANt have natural colors and shadows consistent with the faces. Besides, images generated by EleGANt have the fewest artifacts (4th row) among all of the methods. More samples of comparison are provided in App. 
\ref{sec:add-compare}.

\subsection{Quantitative Comparison. }
We conduct a user study to quantitatively evaluate the generation quality and the transfer precision of different models. For a fair comparison, we compare our EleGANt with the methods whose code and pre-train model are available: BeautyGAN, LADN, PSGAN, and SCGAN. We randomly selected 20 generated images from the test split of the MT dataset. Totally 40 participants were asked to evaluate these samples in three aspects: ``visual quality'', ``detail processing'' (the quality and precision of transferred details), and ``overall performance'' (considering the visual quality, the fidelity of transferred makeup, etc.). They then selected the best one in each aspect. Table \ref{tab:user} demonstrates the results of the user study. Our EleGAnt outperforms other methods in all aspects, especially in synthesizing makeup details.

\begin{table}[!htbp]
\setlength{\belowdisplayskip}{0cm}
\setlength{\abovecaptionskip}{0cm}
\centering
\caption{User study results (ratio ($\%$) selected as the best). ``Quality'', ``Detail'' and ``Overall'' denote the three aspects for evaluation: visual quality, detail processing, and overall performance.}
\label{tab:user}
\begin{tabular}{c|ccccc}
\hline
        & \;BeautyGAN\; & \;LADN\; & \;PSGAN\; & \;SCGAN\; & \;EleGANt (ours)\; \\ \hline
\;Quality\; &  $6.75$ & $1.88$ & $11.13$ & $21.25$ & $\mathbf{59.00}$ \\
\;Detail\; & $5.38$ & $2.75$ & $11.25$ & $14.38$ & $\mathbf{66.25}$ \\ \hline
\;\textbf{Overall}\; &  $4.75$ & $2.75$ & $9.88$ & $20.38$ & $\mathbf{62.25}$ \\ \hline
\end{tabular}
\vskip -0.7cm
\end{table}

\subsection{Controllable Makeup Transfer}\label{sec:control}
Since the makeup feature maps $\Gamma_H$ and $\Gamma_L$ spatially correspond to the source face and are combined with the source feature maps by element-wise multiplication, controllability can be achieved by interpolating those makeup feature maps.

\subsubsection{Partial and Interpolated Makeup Transfer.}
\begin{figure}[t]
    \setlength{\abovecaptionskip}{-0.1cm}
    \setlength{\belowcaptionskip}{0cm}
    \centering
    \includegraphics[width=1.0\textwidth]{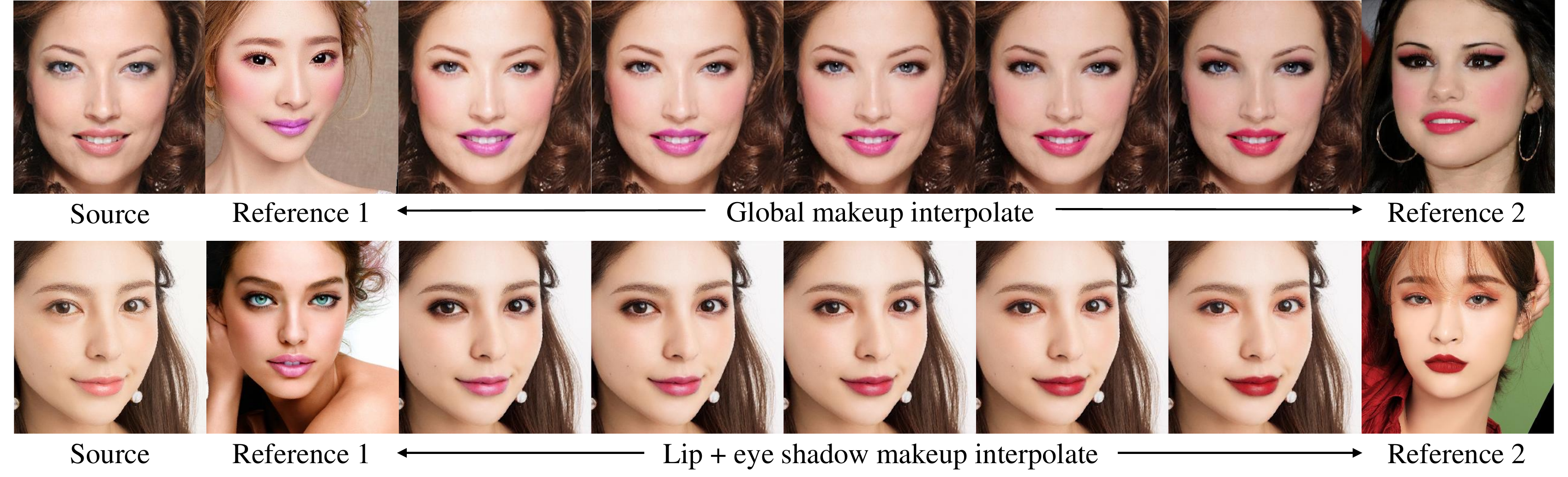}
    \caption{Partial and interpolated makeup transfer. The first row applies a global transfer. The second row only transfers the lipsticks and eye shadows of the two references. }
    \label{fig:control}
\end{figure}

 Transferring the makeup in predefined parts of the face, for instance, lip and skin, can be realized automatically by masking makeup feature maps using face parsing results. Specifically, let $x$ denote the source image, $y_i$ denote the reference image we would like to transfer for part $i$, and $M_i^x\in [0,1]^{H\times W}$ is the corresponding parsing mask, the partial transferred feature maps are calculated as: 

\begin{equation}
\begin{aligned}
\Gamma_* &= M_i^x\odot \Gamma_*^{y_i} + (1- M_i^x) \odot \Gamma_*^x
\end{aligned}
\end{equation}
where $*\in \{H,L\}$, $\Gamma_*^y$ is the makeup feature map with $x$ being the source and $y$ being the reference, the mask is expanded along the channel dimension, and $\odot$ denotes the Hadamard product. Interpolating the makeup feature maps can control the shade of makeup or fuse makeup from multiple references. Given a source image $x$, two reference images $y_1,y_2$, and a coefficient $\alpha^S\in[0,1]$, we first get the makeup feature maps of $y_1$ and $y_2$ w.r.t. $x$, and interpolate them by 
\begin{equation}
\begin{aligned}
\Gamma_*&= \alpha^S \Gamma_*^{y_1} + (1-\alpha^S) \Gamma_*^{y_2}\ ,\ *\in\{H,L\}
\end{aligned}
\end{equation}
 If we want to adjust the shade of makeup using a single reference style, just simply set $y_2=x$, then $\alpha^S$ indicates the intensity. We also perform partial and interpolated makeup transfer simultaneously by leveraging both area masks and shade intensity, and the results are shown in Fig. \ref{fig:control}. 

\begin{figure}[t]
    \setlength{\abovedisplayskip}{0cm}
    \setlength{\abovecaptionskip}{-0.1cm}
    \setlength{\belowcaptionskip}{-0.2cm}
    \centering
    \includegraphics[width=1.0\textwidth]{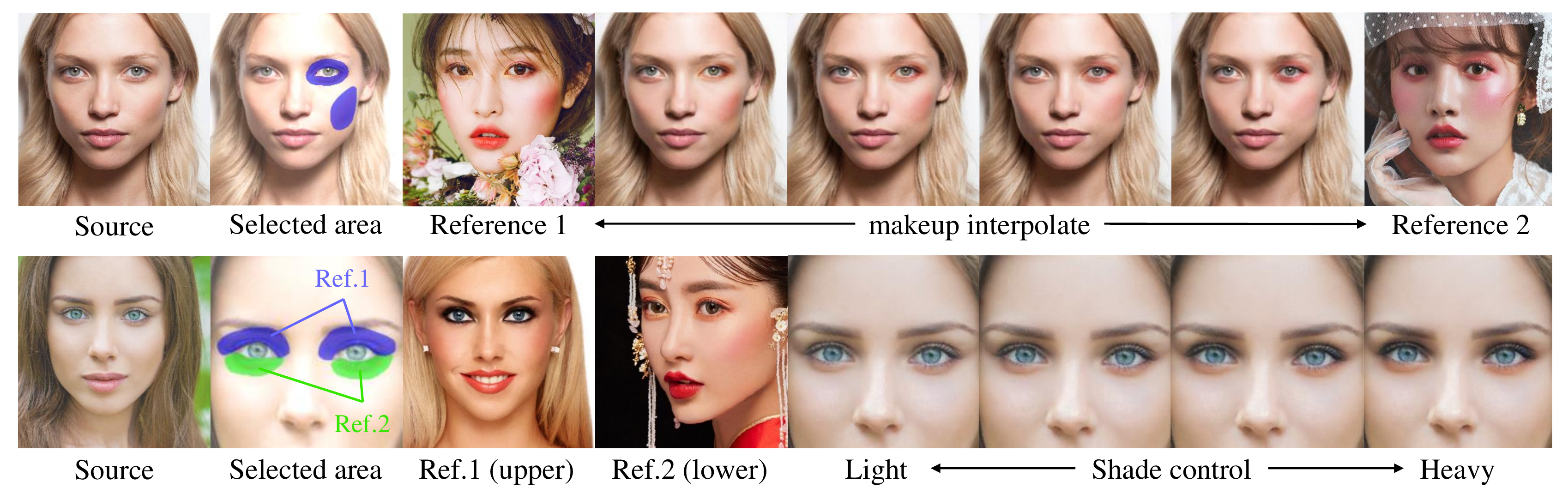}    \caption{Customized local editing. In the first row, we select the areas around the eyes and cheek to adjust the eye shadow and blush. In the second row, we select the areas of upper and lower eye shadow and assign them with different references.}
    \label{fig:paint}
\end{figure}
\subsubsection{Customized Local Editing.}
The general controls can be formulated as: given a set of $k$ reference images $\{y_i\}_{i=1}^k$ with corresponding masks $\{M_i^x\}_{i=1}^k$ for the area to apply makeup and coefficients $\{\alpha^S_i\}_{i=1}^k$ to specify the shade, the fused makeup feature maps are computed by 
\begin{equation}
\label{eq:free}
\begin{aligned}
\Gamma_* &= \sum_{i=1}^k \alpha^S_i M_i^x\odot \Gamma_*^{y_i} + \left(1-\sum_{i=1}^k \alpha^S_i M_i^x\right)\odot\Gamma_*^x\ ,\ *\in\{H,L\}
\end{aligned}
\end{equation}
Unlike 
SCGAN \cite{SCGAN} that restrict the areas for partial transfer to skin, lip and eye shadow, our EleGANt is more interactive and flexible. In Eq. \ref{eq:free}, $M_i^x$ can be the mask of any arbitrary area, and it has the same size as the makeup feature map. Therefore, we can specify customized regions to edit their makeup styles and shades, and the paint-board for region selection is of the same resolution as the high-res makeup feature map $\Gamma_H$. Though the masks need to be down-sampled when applied to the low-res $\Gamma_L$, the makeup details are dominantly determined by $\Gamma_H$, whose high resolution guarantees the precision of the controls. Fig. \ref{fig:paint} illustrates customized local editing of makeup style and shade.

\subsection{Ablation Study}
\label{sec:ablation}
\begin{figure}[t]
    \setlength{\abovecaptionskip}{-0.1cm}
    \setlength{\belowcaptionskip}{0.1cm}
    \centering
    \includegraphics[width=1.0\textwidth]{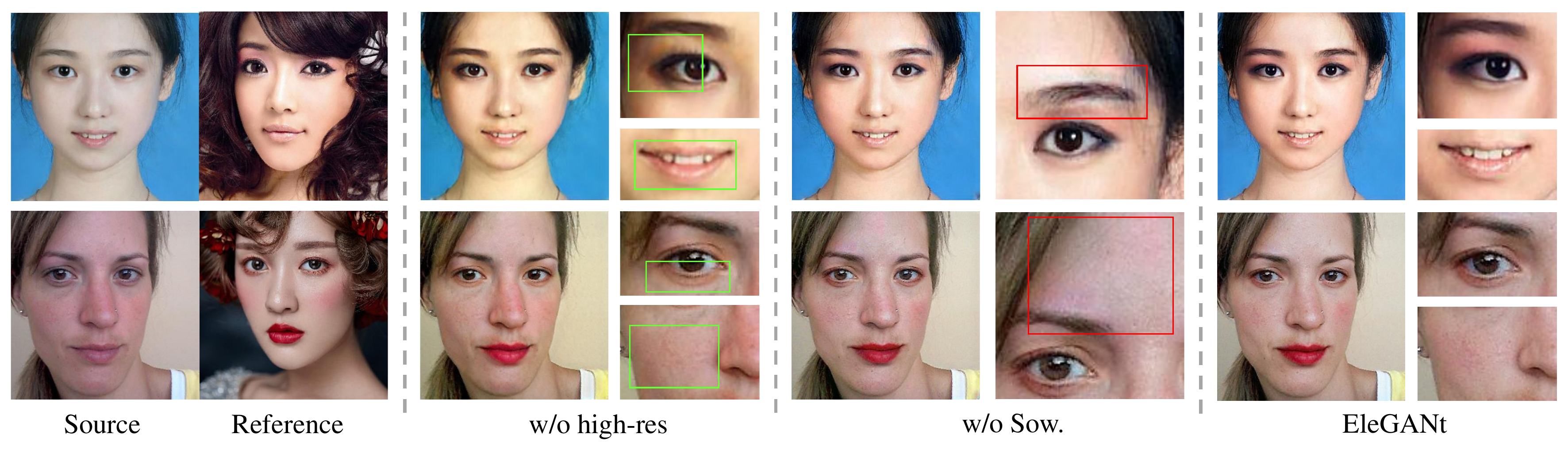}
    \caption{Ablation study of architecture design. ``w/o Sow.'' denotes performing attention in non-overlapped windows instead of shifted overlapped windows. Unnatural colors are marked by red boxes. Detail missing and blurs are marked by green boxes.}
    \label{fig-abla-arch}
\end{figure}
\begin{figure}[t]
    \setlength{\abovecaptionskip}{0.1cm}
    \setlength{\belowcaptionskip}{-0.2cm}
    \centering
    \includegraphics[width=0.92\textwidth]{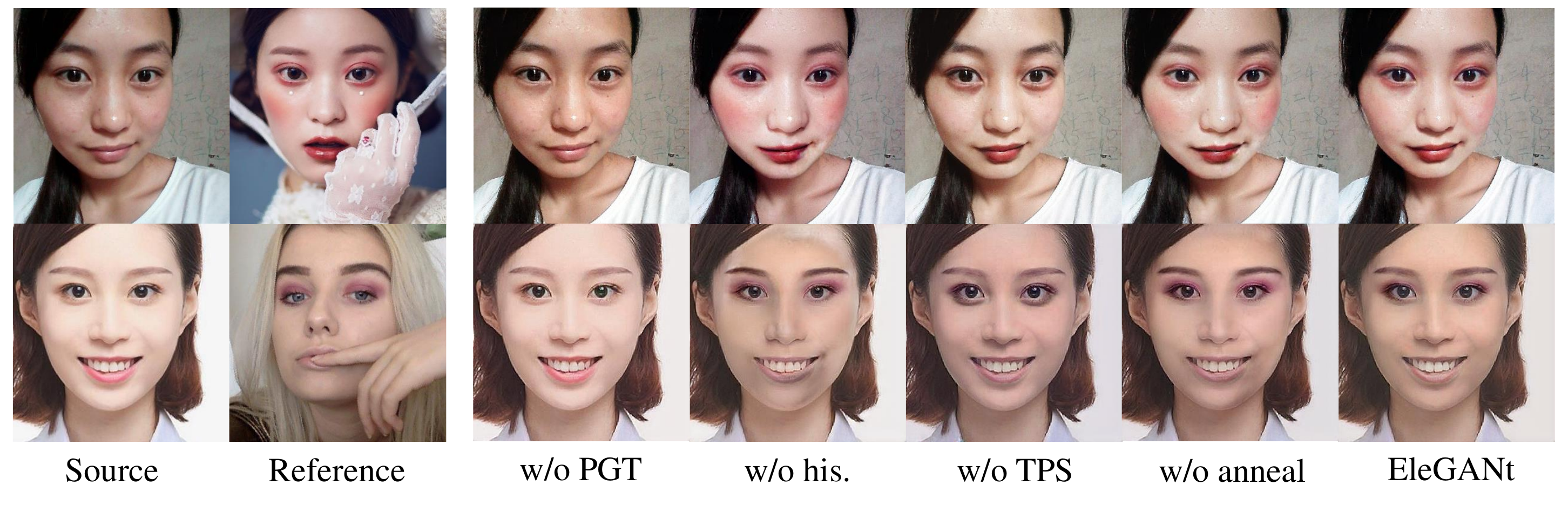}
    \caption{Ablation study of PGT generation. We train the network with different PGT settings and compare the final results generated by the network. ``w/o his.'' denotes generating PGT without histogram matching. ``w/o anneal'' indicates using fixed blending factors rather than annealing during the training process.}
    \label{fig-abla-pgt}
\end{figure}

\subsubsection{Architecture Design.} 
Our EleGANt utilizes high-resolution feature maps to preserve high-frequency information as well as attention in shifted overlapped windows which reduces the complexity and avoids the boundary issue. Here we conduct an ablation study to evaluate their effectiveness. As shown in Fig. \ref{fig-abla-arch}, without the high-res feature maps, detailed facial attributes, e.g., eye shadows of the reference (the 1st and 2nd rows) and freckles of the source (2nd row), are lost or smoothed during the transfer. When replacing Sow-attention with attention in non-overlapped windows, the outputs are discontinuous on the boundaries of the windows. Fig. \ref{fig-abla-arch} shows that there are color blocks nearby the eyebrow (1st row) and on the forehead (2nd row). 

\subsubsection{PGT Generation.} To confirm our proposed strategy for pseudo ground truth generation, we conduct ablation studies on all techniques we employ. Qualitative comparisons are demonstrated in Fig. \ref{fig-abla-pgt}. The model will only yield images with general instead of particular makeups if we do not use PGT for supervision. Spatial information cannot be injected into PGT without TPS transformation, so the model fails to transfer makeup details (e.g., the eye shadows are missing in the first row and inaccurate in the second row.) due to the lack of corresponding supervision. Without histogram matching or annealing, artifacts introduced by TPS warping and image blending remain in the PGT, which are then learned by the model (e.g., stitching traces nearby the lip of the first row and on the forehead of the second row). Besides, the model learns to directly copy the shadows of the reference to the source face when only guided by TPS-warped images.

\section{Conclusion}
In this paper, we emphasize that makeup transfer is beyond transferring color distributions. 
We propose Exquisite and locally editable GAN for makeup transfer (EleGANt) to improve the synthesis of details and step towards more flexible controls. It utilizes high-res feature maps to preserve high-frequency attributes, and a novel Sow-Attention Module performs attention within shifted overlapped to reduce the computational cost. The model is trained with a newly designed objective that leverages both color and spatial information. Besides partial and interpolated makeup transfer, it is the first to achieve customized local makeup editing within arbitrary regions. Extensive experiments demonstrate the superiority of EleGANt compared with existing approaches. It can generate realistic images with exquisite details, despite various facial poses and complex makeup styles. 
Besides, we believe that the attention scheme with shifted overlapped windowing (Sow-Attention) would be helpful for other tasks and networks. 

Although our model succeeds in daily makeup transfer, it fails in some cases such as extreme makeup. The MT dataset \cite{BeautyGAN} we currently use has limited resolution and diversity of makeup styles and skin tones. This may be addressed if more data is available and we leave this for future work.

\bigskip\noindent\textbf{Acknowledgements.} This work is supported by the Ministry of Science and Technology of the People´s Republic of China, the 2030 Innovation Megaprojects ``Program on New Generation Artificial Intelligence'' (Grant No. 2021AAA0150000). This work is also supported by a grant from the Guoqiang Institute, Tsinghua University. Thanks to Steve Lin for his pre-reading and constructive suggestions.
%
%
\bibliographystyle{splncs04}
\bibliography{egbib}

\begin{thebibliography}{10}
\providecommand{\url}[1]{\texttt{#1}}
\providecommand{\urlprefix}{URL }
\providecommand{\doi}[1]{https://doi.org/#1}

\bibitem{style-3}
An, J., Xiong, H., Huan, J., Luo, J.: Ultrafast photorealistic style transfer
  via neural architecture search. In: Proceedings of the AAAI Conference on
  Artificial Intelligence. vol.~34, pp. 10443--10450 (2020)

\bibitem{FDIT}
Cai, M., Zhang, H., Huang, H., Geng, Q., Li, Y., Huang, G.: Frequency domain
  image translation: More photo-realistic, better identity-preserving. In:
  Proceedings of the IEEE/CVF International Conference on Computer Vision
  (ICCV). pp. 13930--13940 (2021)

\bibitem{PairedCycleGAN}
Chang, H., Lu, J., Yu, F., Finkelstein, A.: Pairedcyclegan: Asymmetric style
  transfer for applying and removing makeup. In: Proceedings of the IEEE/CVF
  Conference on Computer Vision and Pattern Recognition (CVPR). pp. 40--48
  (2018)

\bibitem{R2L}
Chen, C.F., Panda, R., Fan, Q.: Regionvit: Regional-to-local attention for
  vision transformers. In: Proceedings of the International Conference on
  Learning Representations (ICLR) (2022)

\bibitem{BeautyGlow}
Chen, H.J., Hui, K.M., Wang, S.Y., Tsao, L.W., Shuai, H.H., Cheng, W.H.:
  Beautyglow: On-demand makeup transfer framework with reversible generative
  network. In: Proceedings of the IEEE/CVF Conference on Computer Vision and
  Pattern Recognition (CVPR). pp. 10042--10050 (2019)

\bibitem{StarGAN}
Choi, Y., Uh, Y., Yoo, J., Ha, J.W.: Stargan v2: Diverse image synthesis for
  multiple domains. In: Proceedings of the IEEE/CVF Conference on Computer
  Vision and Pattern Recognition (CVPR). pp. 8188--8197 (2020)

\bibitem{Twins}
Chu, X., Tian, Z., Wang, Y., Zhang, B., Ren, H., Wei, X., Xia, H., Shen, C.:
  Twins: Revisiting the design of spatial attention in vision transformers. In:
  Proceedings of the International Conference on Neural Information Processing
  Systems (NIPS). pp. 9355--9366 (2021)

\bibitem{SCGAN}
Deng, H., Han, C., Cai, H., Han, G., He, S.: Spatially-invariant style-codes
  controlled makeup transfer. In: Proceedings of the IEEE/CVF Conference on
  Computer Vision and Pattern Recognition (CVPR). pp. 6549--6557 (2021)

\bibitem{ViT}
Dosovitskiy, A., Beyer, L., Kolesnikov, A., Weissenborn, D., Zhai, X.,
  Unterthiner, T., Dehghani, M., Minderer, M., Heigold, G., Gelly, S.,
  Uszkoreit, J., Houlsby, N.: An image is worth 16x16 words: Transformers for
  image recognition at scale. In: Proceedings of the International Conference
  on Learning Representations (ICLR) (2021)

\bibitem{NeuralTransfer}
Gatys, L.A., Ecker, A.S., Bethge, M.: Image style transfer using convolutional
  neural networks. In: Proceedings of the IEEE Conference on Computer Vision
  and Pattern Recognition (CVPR). pp. 2414--2423 (2016)

\bibitem{GAN}
Goodfellow, I., Pouget-Abadie, J., Mirza, M., Xu, B., Warde-Farley, D., Ozair,
  S., Courville, A., Bengio, Y.: Generative adversarial nets. In: Proceedings
  of the International Conference on Neural Information Processing Systems
  (NIPS) (2014)

\bibitem{LADN}
Gu, Q., Wang, G., Chiu, M.T., Tai, Y.W., Tang, C.K.: Ladn: Local adversarial
  disentangling network for facial makeup and de-makeup. In: Proceedings of the
  IEEE/CVF International Conference on Computer Vision (ICCV). pp. 10481--10490
  (2019)

\bibitem{Example-2}
Guo, D., Sim, T.: Digital face makeup by example. In: Proceedings of the IEEE
  Conference on Computer Vision and Pattern Recognition (CVPR). pp. 73--79
  (2009)

\bibitem{PiT}
Heo, B., Yun, S., Han, D., Chun, S., Choe, J., Oh, S.J.: Rethinking spatial
  dimensions of vision transformers. In: Proceedings of the IEEE/CVF
  International Conference on Computer Vision (ICCV). pp. 11936--11945 (2021)

\bibitem{axial-1}
Ho, J., Kalchbrenner, N., Weissenborn, D., Salimans, T.: Axial attention in
  multidimensional transformers. arXiv preprint arXiv:1912.12180  (2019)

\bibitem{AdaIN}
Huang, X., Belongie, S.: Arbitrary style transfer in real-time with adaptive
  instance normalization. In: Proceedings of the IEEE International Conference
  on Computer Vision (ICCV). pp. 1501--1510 (2017)

\bibitem{Pix2Pix}
Isola, P., Zhu, J.Y., Zhou, T., Efros, A.A.: Image-to-image translation with
  conditional adversarial networks. In: Proceedings of the IEEE Conference on
  Computer Vision and Pattern Recognition (CVPR). pp. 1125--1134 (2017)

\bibitem{STN}
Jaderberg, M., Simonyan, K., Zisserman, A., Kavukcuoglu, K.: Spatial
  transformer networks. In: Proceedings of the International Conference on
  Neural Information Processing Systems (NIPS). pp. 2017--2025 (2015)

\bibitem{PSGAN}
Jiang, W., Liu, S., Gao, C., Cao, J., He, R., Feng, J., Yan, S.: Psgan: Pose
  and expression robust spatial-aware gan for customizable makeup transfer. In:
  Proceedings of the IEEE/CVF Conference on Computer Vision and Pattern
  Recognition (CVPR). pp. 5194--5202 (2020)

\bibitem{PerceptualLoss}
Johnson, J., Alahi, A., Fei-Fei, L.: Perceptual losses for real-time style
  transfer and super-resolution. In: Proceedings of the European Conference on
  Computer Vision (ECCV). pp. 694--711 (2016)

\bibitem{style-2}
Kim, S.S., Kolkin, N., Salavon, J., Shakhnarovich, G.: Deformable style
  transfer. In: Proceedings of the European Conference on Computer Vision
  (ECCV). pp. 246--261 (2020)

\bibitem{Adam}
Kingma, D.P., Ba, J.: Adam: {A} method for stochastic optimization. In: Bengio,
  Y., LeCun, Y. (eds.) Proceedings of the International Conference on Learning
  Representations (ICLR) (2015)

\bibitem{tradition-2}
Li, C., Zhou, K., Lin, S.: Simulating makeup through physics-based manipulation
  of intrinsic image layers. In: Proceedings of the IEEE Conference on Computer
  Vision and Pattern Recognition (CVPR). pp. 4621--4629 (2015)

\bibitem{BeautyGAN}
Li, T., Qian, R., Dong, C., Liu, S., Yan, Q., Zhu, W., Lin, L.: Beautygan:
  Instance-level facial makeup transfer with deep generative adversarial
  network. In: Proceedings of the 26th ACM International Conference on
  Multimedia. pp. 645--653 (2018)

\bibitem{DIA}
Liao, J., Yao, Y., Yuan, L., Hua, G., Kang, S.B.: Visual attribute transfer
  through deep image analogy. ACM Transactions on Graphics (TOG)
  \textbf{36}(4),  1--15 (2017)

\bibitem{tradition-3}
Liu, L., Xing, J., Liu, S., Xu, H., Zhou, X., Yan, S.: Wow! you are so
  beautiful today! ACM Transactions on Multimedia Computing, Communications,
  and Applications (TOMM)  \textbf{11}(1s),  1--22 (2014)

\bibitem{Swin}
Liu, Z., Lin, Y., Cao, Y., Hu, H., Wei, Y., Zhang, Z., Lin, S., Guo, B.: Swin
  transformer: Hierarchical vision transformer using shifted windows. In:
  Proceedings of the IEEE/CVF International Conference on Computer Vision
  (ICCV). pp. 10012--10022 (2021)

\bibitem{phototransfer}
Luan, F., Paris, S., Shechtman, E., Bala, K.: Deep photo style transfer. In:
  Proceedings of the IEEE Conference on Computer Vision and Pattern Recognition
  (CVPR). pp. 4990--4998 (2017)

\bibitem{local-attn-1}
Parmar, N., Vaswani, A., Uszkoreit, J., Kaiser, L., Shazeer, N., Ku, A., Tran,
  D.: Image transformer. In: Proceedings of the International Conference on
  Machine Learning (ICML). pp. 4055--4064 (2018)

\bibitem{block-1}
Rakhimov, R., Volkhonskiy, D., Artemov, A., Zorin, D., Burnaev, E.: Latent
  video transformer. In: Proceedings of the 16th International Joint Conference
  on Computer Vision, Imaging and Computer Graphics Theory and Applications
  (VISIGRAPP). pp. 101--112 (2021)

\bibitem{ASTER}
Shi, B., Yang, M., Wang, X., Lyu, P., Yao, C., Bai, X.: Aster: An attentional
  scene text recognizer with flexible rectification. IEEE Transactions on
  Pattern Analysis and Machine Intelligence  \textbf{41}(9),  2035--2048 (2018)

\bibitem{Example-1}
Tong, W.S., Tang, C.K., Brown, M.S., Xu, Y.Q.: Example-based cosmetic transfer.
  In: Proceedings of the 15th Pacific Conference on Computer Graphics and
  Applications (PG). pp. 211--218 (2007)

\bibitem{Transformer}
Vaswani, A., Shazeer, N., Parmar, N., Uszkoreit, J., Jones, L., Gomez, A.N.,
  Kaiser, {\L}., Polosukhin, I.: Attention is all you need. In: Proceedings of
  the International Conference on Neural Information Processing Systems (NIPS).
  pp. 6000--6010 (2017)

\bibitem{FAT}
Wan, Z., Chen, H., An, J., Jiang, W., Yao, C., Luo, J.: Facial attribute
  transformers for precise and robust makeup transfer. In: Proceedings of the
  IEEE/CVF Winter Conference on Applications of Computer Vision (WACV). pp.
  1717--1726 (2022)

\bibitem{style-1}
Wang, H., Li, Y., Wang, Y., Hu, H., Yang, M.H.: Collaborative distillation for
  ultra-resolution universal style transfer. In: Proceedings of the IEEE/CVF
  Conference on Computer Vision and Pattern Recognition (CVPR). pp. 1860--1869
  (2020)

\bibitem{PVT}
Wang, W., Xie, E., Li, X., Fan, D.P., Song, K., Liang, D., Lu, T., Luo, P.,
  Shao, L.: Pyramid vision transformer: A versatile backbone for dense
  prediction without convolutions. In: Proceedings of the IEEE/CVF
  International Conference on Computer Vision (ICCV). pp. 568--578 (2021)

\bibitem{vattn-2}
Wang, X., Girshick, R., Gupta, A., He, K.: Non-local neural networks. In:
  Proceedings of the IEEE Conference on Computer Vision and Pattern Recognition
  (CVPR). pp. 7794--7803 (2018)

\bibitem{block-2}
Weissenborn, D., T{\"{a}}ckstr{\"{o}}m, O., Uszkoreit, J.: Scaling
  autoregressive video models. In: Proceedings of the International Conference
  on Learning Representations (ICLR) (2020)

\bibitem{GODIVA}
Wu, C., Huang, L., Zhang, Q., Li, B., Ji, L., Yang, F., Sapiro, G., Duan, N.:
  Godiva: Generating open-domain videos from natural descriptions. arXiv
  preprint arXiv:2104.14806  (2021)

\bibitem{NUWA}
Wu, C., Liang, J., Ji, L., Yang, F., Fang, Y., Jiang, D., Duan, N.:
  N$\backslash$" uwa: Visual synthesis pre-training for neural visual world
  creation. arXiv preprint arXiv:2111.12417  (2021)

\bibitem{vattn-1}
Xu, K., Ba, J., Kiros, R., Cho, K., Courville, A., Salakhudinov, R., Zemel, R.,
  Bengio, Y.: Show, attend and tell: Neural image caption generation with
  visual attention. In: Proceedings of the International Conference on Machine
  Learning (ICML). pp. 2048--2057 (2015)

\bibitem{tradition-1}
Xu, L., Du, Y., Zhang, Y.: An automatic framework for example-based virtual
  makeup. In: Proceedings of the IEEE International Conference on Image
  Processing (ICIP). pp. 3206--3210 (2013)

\bibitem{T2T}
Yuan, L., Chen, Y., Wang, T., Yu, W., Shi, Y., Jiang, Z.H., Tay, F.E., Feng,
  J., Yan, S.: Tokens-to-token vit: Training vision transformers from scratch
  on imagenet. In: Proceedings of the IEEE/CVF International Conference on
  Computer Vision (ICCV). pp. 558--567 (2021)

\bibitem{landmark}
Zhang, K., Zhang, Z., Li, Z., Qiao, Y.: Joint face detection and alignment
  using multitask cascaded convolutional networks. IEEE Signal Processing
  Letters  \textbf{23}(10),  1499--1503 (2016)

\bibitem{ViL}
Zhang, P., Dai, X., Yang, J., Xiao, B., Yuan, L., Zhang, L., Gao, J.:
  Multi-scale vision longformer: A new vision transformer for high-resolution
  image encoding. In: Proceedings of the IEEE/CVF International Conference on
  Computer Vision (ICCV). pp. 2998--3008 (2021)

\bibitem{CycleGAN}
Zhu, J.Y., Park, T., Isola, P., Efros, A.A.: Unpaired image-to-image
  translation using cycle-consistent adversarial networks. In: Proceedings of
  the IEEE International Conference on Computer Vision (ICCV). pp. 2223--2232
  (2017)

\end{thebibliography}


\appendix
\section{Landmark Embedding}
We adopt the Landmark Embedding proposed by \cite{FAT} to introduce spatial information into attention. It utilizes facial landmarks as anchor points to represent relative positions. Given $N$ landmark points $\{L_n\}_{n=1}^N$ of the facial image, each pixel $x_i$ on the image is assigned with a vector $\mathbf{p}_i\in\mathbb{R}^{2N}$ as positional embedding, which is computed by its relative positions to those landmark points:
\begin{equation}
\begin{aligned}
    \mathbf{p}_i^{(2n)}&=\mathrm{x}(x_i) - \mathrm{x}(L_n)\\
    \mathbf{p}_i^{(2n+1)}&=\mathrm{y}(x_i) - \mathrm{y}(L_n) 
\end{aligned}
\ ,\ n=1,\dots, N 
\end{equation}
where $\mathrm{x}(\cdot)$ and $\mathrm{y}(\cdot)$ denote the x-coordinate and y-coordinate of a point respectively. The spatial feature $\mathbf{p}_i$ is then normalized by its 2-norm to ensure size-invariance, and is concatenated with the visual features at pixel $x_i$.

\section{TPS Transformation}
\label{sec:tps}
We adopt STN \cite{STN} that uses a grid generator to compute a sampling grid $\mathcal{P} = \{p_i\}$ on an image to form a transformation. A 2D TPS transformation with $N$ control points $\mathbf{C},\mathbf{C'}\in\mathbb{R}^{2\times N}$ is parameterized by a $2\times(N + 3)$ matrix:
\begin{align}
    \mathbf{T}=\left[
    \begin{matrix}
    a_0 & a_1 & a_2 & \mathbf{u}\\
    b_0 & b_1 & b_2 & \mathbf{v}\\
    \end{matrix}\right]
\end{align}
where $\mathbf{u},\mathbf{v}\in\mathbb{R}^{1\times N}$. We follow the formulation in \cite{ASTER} to describe the grid computation of 2D TPS. For a point $\mathbf{p}\in\mathbb{R}^{1\times2}$, its sampling point $\mathbf{p}'$ is computed by a linear projection: 
\begin{align}
    \mathbf{p}' = \mathbf{T}\left[
    \begin{matrix}
    1\\
    \mathbf{p}\\
    \phi (||\mathbf{p}-\mathbf{c}_1||)\\
    \cdots\\
    \phi (||\mathbf{p}-\mathbf{c}_N||)\\
    \end{matrix}\right]
\end{align}
where $\phi(r)=r^2\log r$ is the radial basis kernel applied to the Euclidean distance between $\mathbf{p}$ and the control points $\mathbf{C}$. The coefficients of TPS are obtained by solving a linear system involving $N$ correspondences between $\mathbf{C}$ and $\mathbf{C}'$:
\begin{align}
    \mathbf{c}_i' = \mathbf{T}\left[
    \begin{matrix}
    1\\
    \mathbf{c}_i\\
    \phi (||\mathbf{c}_i-\mathbf{c}_1||)\\
    \cdots\\
    \phi (||\mathbf{c}_i-\mathbf{c}_N||)\\
    \end{matrix}\right],\ i=1,\dots N
\end{align}
subject to the following boundary conditions:
\begin{align}
    \begin{split}
        0&=\mathbf{u}\mathbf{1}\\
        0&=\mathbf{v}\mathbf{1}\\
        0&=\mathbf{u}\mathbf{C}^T_x\\
        0&=\mathbf{v}\mathbf{C}^T_y\\
    \end{split}
\end{align}
where $\mathbf{C}_x$ and $\mathbf{C}_y$ are the $x$ and $y$ coordinates of $\mathbf{C}$, respectively. $\mathbf{T}$ has a closed-form solution in matrix form:
\begin{align}
    \begin{split}
        \mathbf{T}=\left[\begin{matrix}
        C' & \mathbf{0}^{2\times 3}\\
        \end{matrix}\right]\mathbf{\Delta}_C^{-1}\\
        \mathbf{\Delta}_C = \left[
        \begin{matrix}
        \mathbf{1}^{1\times N} & \mathbf{0} & \mathbf{0}\\
        \mathbf{C} & \mathbf{0} & \mathbf{0}\\
        \hat{\mathbf{C}} & \mathbf{1}^{N\times 1} & \mathbf{C}^T\\
        \end{matrix}
        \right]
    \end{split}
\end{align}
where $\hat{\mathbf{C}}\in\mathbb{R}^{N\times N}$ is a square matrix comprising $\hat{\mathbf{C}}_{i,j}=\phi(||\mathbf{c}_i-\mathbf{c}_j||)$.

\section{Full Objective} \label{sec:objective}
Given the non-makeup domain $X$ and the makeup domain $Y$, our proposed EleGANt learns the mapping bidirectionally between these two domains. Let $x\in X$ and $y\in Y$ denote the source image and the reference makeup image, respectively. $\hat{x}=\mathcal{G}(x,y)$ is the transferred result with the
makeup style of $y$ and the facial identity of $x$.

\subsubsection{Adversarial Loss.} 
Adversarial loss \cite{GAN} is introduced to guide realistic generation. We apply two discriminators $\mathcal{D}_{X}$ and $\mathcal{D}_{Y}$ to discriminate between real and generated images in the domain $X$ and $Y$ respectively.
The adversarial loss $L_{\mathcal{G}}^{adv}$ for generator and $L_{\mathcal{D}}^{adv}$ for discriminator are defined as

\begin{equation}
\begin{aligned}
L_{\mathcal{G}}^{adv} =&-\mathbb{E}_{x \sim {X}, y \sim {Y}}\left[\log \left(\mathcal{D}_{X}(\mathcal{G}(y, x))\right)\right] \\
&-\mathbb{E}_{x \sim {X}, y \sim {Y}}\left[\log \left(\mathcal{D}_{Y}(\mathcal{G}(x, y))\right)\right]
\end{aligned}
\end{equation}

\begin{equation}
\begin{aligned}
L_{\mathcal{D}}^{a d v} =&-\mathbb{E}_{x \sim {X}}\left[\log \mathcal{D}_{X}(x)\right]-\mathbb{E}_{y \sim {Y}}\left[\log \mathcal{D}_{Y}(y)\right] \\
&-\mathbb{E}_{x \sim {X}, y \sim {Y}}\left[\log \left(1-\mathcal{D}_{X}(\mathcal{G}(y, x))\right)\right] \\
&-\mathbb{E}_{x \sim {X}, y \sim {Y}}\left[\log \left(1-\mathcal{D}_{Y}(\mathcal{G}(x, y))\right)\right]
\end{aligned}
\end{equation}

\subsubsection{Cycle Consistency Loss.} 
We use cycle consistency loss \cite{CycleGAN} for unsupervised learning with unpaired images. The cycle consistency loss $L_{\mathcal{G}}^{c y c}$ is defined as the $L_1$-distance between the original image and reconstructed image:

\begin{equation}
\begin{aligned}
L_{\mathcal{G}}^{c y c} &=\mathbb{E}_{x \sim {X}, y \sim {Y}}\left[\|\mathcal{G}(\mathcal{G}(x, y), x)-x\|_{1}\right] \\
&+\mathbb{E}_{x \sim {X}, y \sim {Y}}\left[\|\mathcal{G}(\mathcal{G}(y, x), y)-y\|_{1}\right]
\end{aligned}
\end{equation}

\subsubsection{Perceptual Loss.} 
To guarantee that the personal identity of the source image is preserved in the transferred image, we utilize perceptual loss \cite{PerceptualLoss} to maintain face identity. $L_{\mathcal{G}}^{per}$ is defined as

\begin{equation}
\begin{aligned}
L_{\mathcal{G}}^{p e r} &=\mathbb{E}_{x \sim {X}, y \sim {Y}}\left[\left\|F_{l}(\mathcal{G}(x, y))-F_{l}(x)\right\|_{2}\right] \\
&+\mathbb{E}_{x \sim {X}, y \sim {Y}}\left[\left\|F_{l}(\mathcal{G}(y, x))-F_{l}(y)\right\|_{2}\right]
\end{aligned}
\end{equation}
where $F_{l}(\cdot)$ is the $l$-th layer output of the pre-trained VGG-16 model.

\subsubsection{Makeup Loss.} To provide guidance for transferring specific makeup styles, we introduce makeup loss as extra supervision. $L_{\mathcal{G}}^{m a k e} $ is defined as the $L_2$-distance between the transferred image and the pseudo ground truth (PGT) generated by our proposed strategy AC-PGT:

\begin{equation}
\begin{aligned}
L_{\mathcal{G}}^{m a k e} &=\|\mathcal{G}(x, y)-P G T(x, y)\|_{1} +\|\mathcal{G}(y, x)-P G T(y, x)\|_{1}
\end{aligned}
\end{equation}

\subsubsection{Total Loss.} The total loss can be expressed as:
\begin{equation}
\begin{aligned}
    L_{\mathcal{G}}&=\lambda_{a d v} L_{\mathcal{G}}^{a d v} +\lambda_{c y c} L_{\mathcal{G}}^{c y c}+\lambda_{per} L_{\mathcal{G}}^{per}+\lambda_{make} L_{\mathcal{G}}^{make}\\
    L_{\mathcal{D}} &= \lambda_{a d v} L_{\mathcal{D}}^{a d v}
\end{aligned}
\end{equation}
where $\lambda_{adv}$, $\lambda_{cyc}$, $\lambda_{per}$, $\lambda_{make}$ are trade-off parameters.

\section{Implementation Details} \label{sec:implement}
\subsection{Training Settings}
The optimizer of the generator the discriminator is Adam \cite{Adam} with $\beta_1=0.5$ = 0.5 and $\beta_2 = 0.999$. The learning rate is initially 2e-4 and decreases to 1e-5 by cosine annealing decay. The model are trained for 50 epochs with batch size 1. We extract features from $relu\_4\_1$ layer of pretrained VGG-16 to calculate the perceptual loss. The trade-off parameters in the loss function are set as $\lambda_{adv}=1$, $\lambda_{cyc}=10$, $\lambda_{per}=0.005$, $\lambda_{make}=1$.

\subsection{Network Architecture}
\begin{figure}[t]
    \centering
    \subfigure[Resblock]{
        \includegraphics[width=0.305\textwidth]{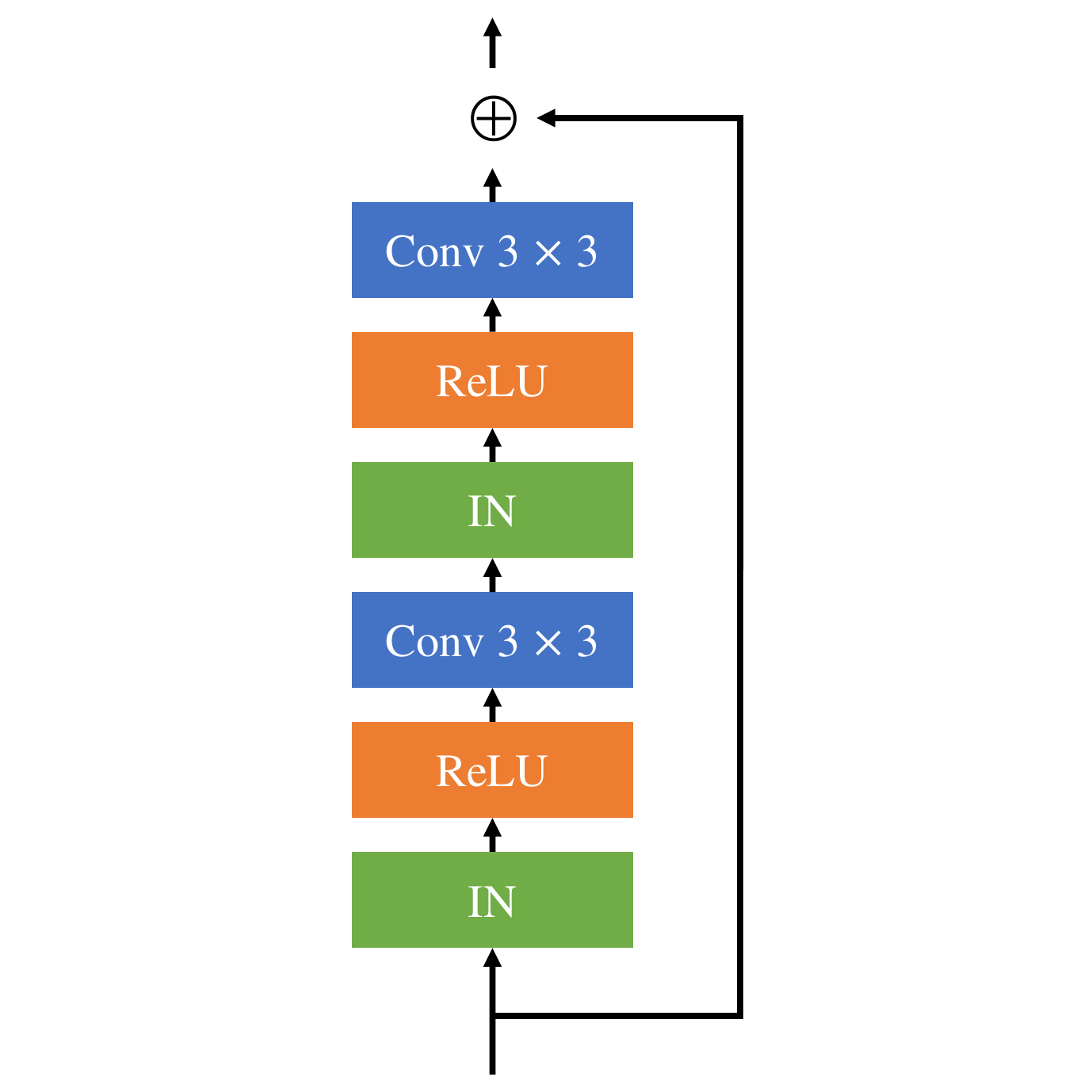}
        \label{Resblock}
    }
    \subfigure[Down-sampling]{
        \includegraphics[width=0.305\textwidth]{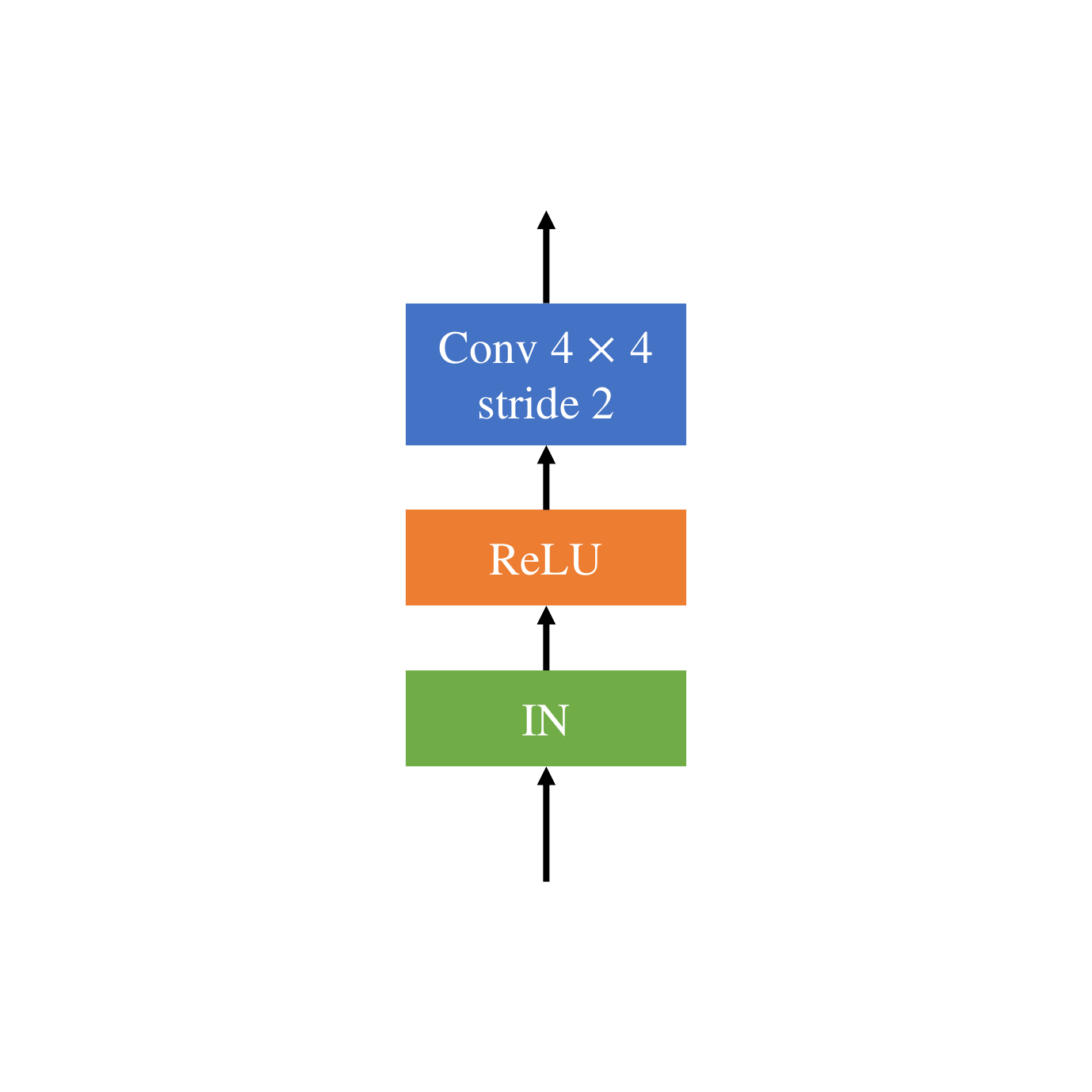}
        \label{Downsample}
    }
    \subfigure[Up-sampling]{
        \includegraphics[width=0.305\textwidth]{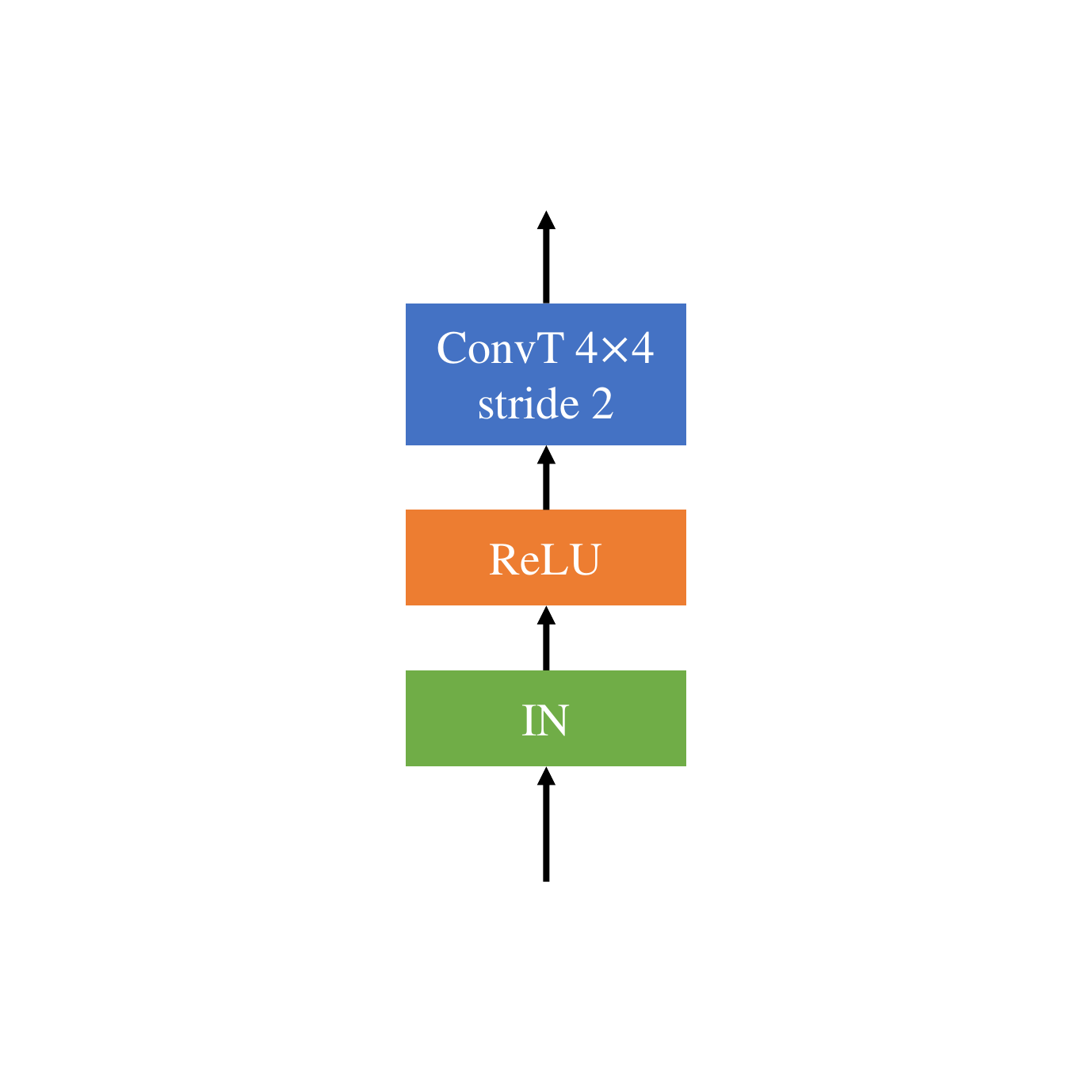}
        \label{Upsample}
    }
    \caption{Basic blocks used in our network. ``IN'' denotes Instance Normalization.}
    \label{fig:block}
\end{figure}

\begin{figure}[t]
    \centering
    \subfigure[FAEnc]{
        \includegraphics[width=0.4\textwidth]{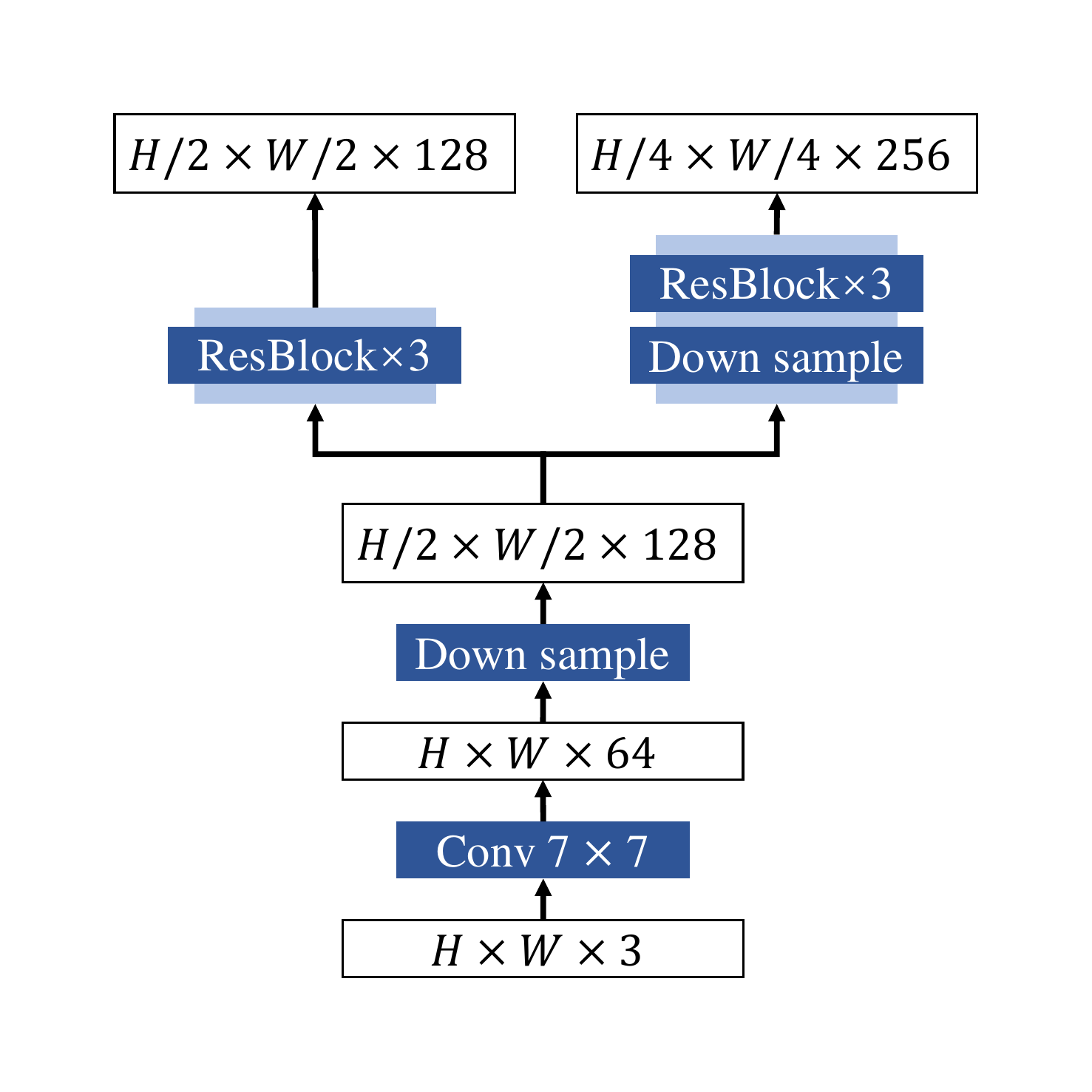}
        \label{FAEnc}
    }
    \subfigure[MADec]{
        \includegraphics[width=0.4\textwidth]{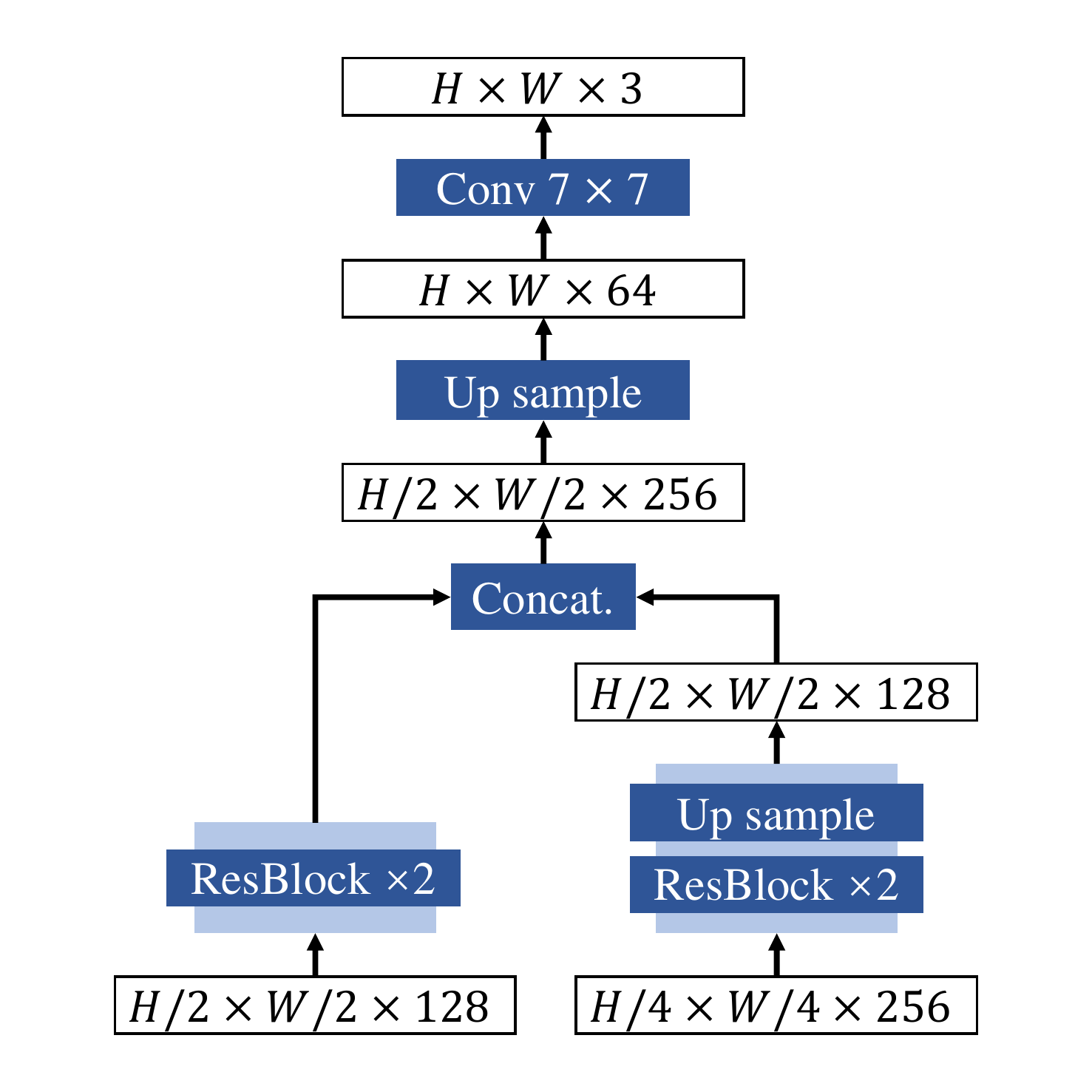}
        \label{MADec}
    }
    \caption{Architecture of FAEnd and MADec in EleGANt. ``Concat.'' denotes concatenation along the channel dimension. ``$\times n$'' indicates a stack of $n$ blocks.}
    \label{fig:arch_app}
\end{figure}

We use three basic blocks to construct the generator of our EleGANt: Resblock, Down-sampling block, and Up-sampling block, whose architectures are shown in Fig. \ref{fig:block}. The architectures of FAEnc and MADec in EleGANt with the shape of corresponding feature maps are illustrated in Fig. \ref{fig:arch_app}. In Makeup Transfer Module (MTM), there is one Attention Module and one Sow-Attention Module, each of which performs cross-attention once as described in Sec. \ref{sec:arch}. We adopt the discriminator in \cite{Pix2Pix} that distinguishes overlapped patches of size $70\times70$ of the image between real and fake.


\subsection{Pseudo Ground Truth}
\label{sec:pgt}
\begin{figure}[!htbp]
    \centering
    \subfigure[$\alpha^D_{skin}$]{
        \includegraphics[width=0.305\textwidth]{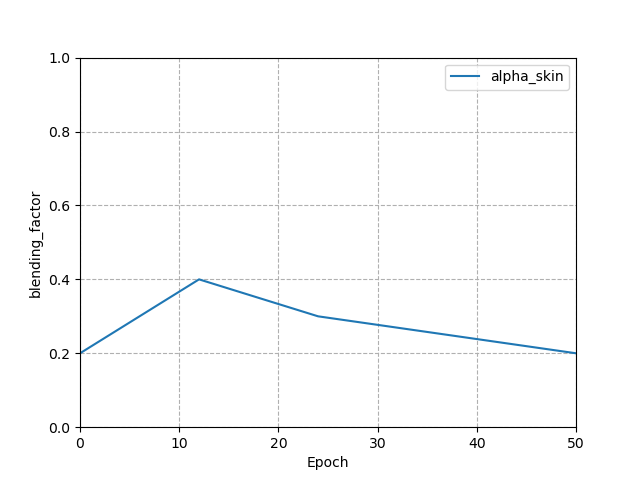}
        \label{alphe_skin}
    }
    \subfigure[$\alpha^D_{lip}$]{
        \includegraphics[width=0.305\textwidth]{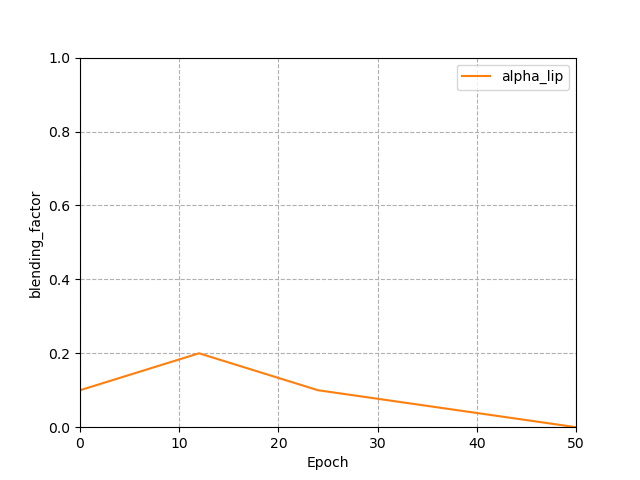}
        \label{alpha_lip}
    }
    \subfigure[$\alpha^D_{eyes}$]{
        \includegraphics[width=0.305\textwidth]{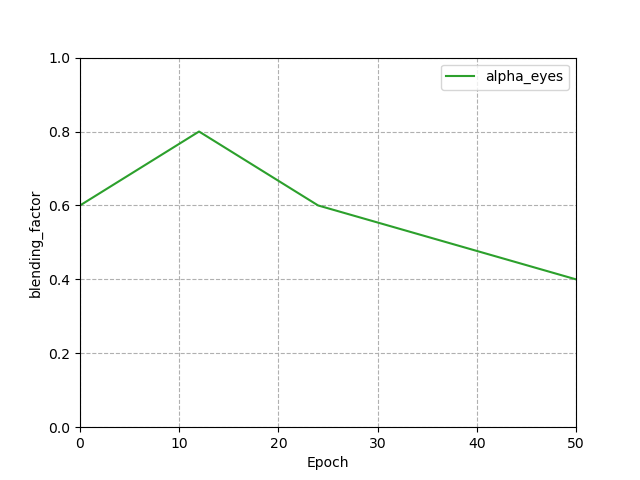}
        \label{alpha_eyes}
    }
    \caption{Annealing functions for the blending factors.}
    \label{fig:alpha}
\end{figure}

Our proposed AC-PGT can be expressed in formula:
\begin{equation}
\begin{aligned}
    PGT(x,y)= \sum_{i\in I} M^x_i\left(\alpha_i^D TPS\left(x,y,C^x_i,C^y_i\right) + \left(1-\alpha_i^D\right) HM(x,y)\right)
\end{aligned}
\end{equation}
where $HM(x,y)$ is the result of histogram matching which has the makeup of $y$ and the identity of $x$, $TPS\left(x,y,C^x_i,C^y_i\right)$ denotes the result of warping $y$ to fit $x$ by TPS using control points $C^x_i$ and $C^y_i$, $M^x_i$ denotes the binary mask, and $\alpha_i^D$ is the blending factor for region $i$ in $I=\{skin, lip, eye shadow\}$. We set the control points as all facial landmarks for the skin region, the landmarks around the eyes for the eye shadow region, and the landmarks on the lip for the lip region. We design annealing functions for blending factors $\alpha^D_{skin}$, $\alpha^D_{eyes}$, and $\alpha^D_{lip}$, which are shown in Fig. \ref{fig:alpha}. In the early stages of training, $\alpha_i^D$ gradually increases to guide the generator to first learn the global transfer for overall colors and then learn the local one for makeup details. In the later stages, we decrease $\alpha^D_i$ to avoid the artifacts in PGT introduced by image warping and stitching being learned. 

\section{Additional Results}
\subsection{User Study}
We have recorded additional information about the participants of the user study. The distribution of their ages and genders are reported in Table. \ref{tab:dis}.
\begin{table}[htbp]
\setlength{\abovecaptionskip}{-0.2cm}
\centering
\caption{Ages and genders of the participants in the user study}
\begin{tabular}{c|cccc}
\hline
Age & [20,30) & [30,40) & [40,50) & [50, 60) \\ \hline
Ratio (\%)  & 62.5 & 27.5 & 7.5 & 2.5 \\ \hline
\end{tabular}
\\
~\\
\begin{tabular}{c|cc}
\hline
Gender & Female & Male \\ \hline
Ratio (\%)  & 55 & 45   \\ \hline
\end{tabular}
\label{tab:dis}
\vskip -0.7cm
\end{table}

\subsection{Qualitative Comparison}
\label{sec:add-compare}
\begin{figure}[!htbp]
    \centering
    \includegraphics[width=1.0\textwidth]{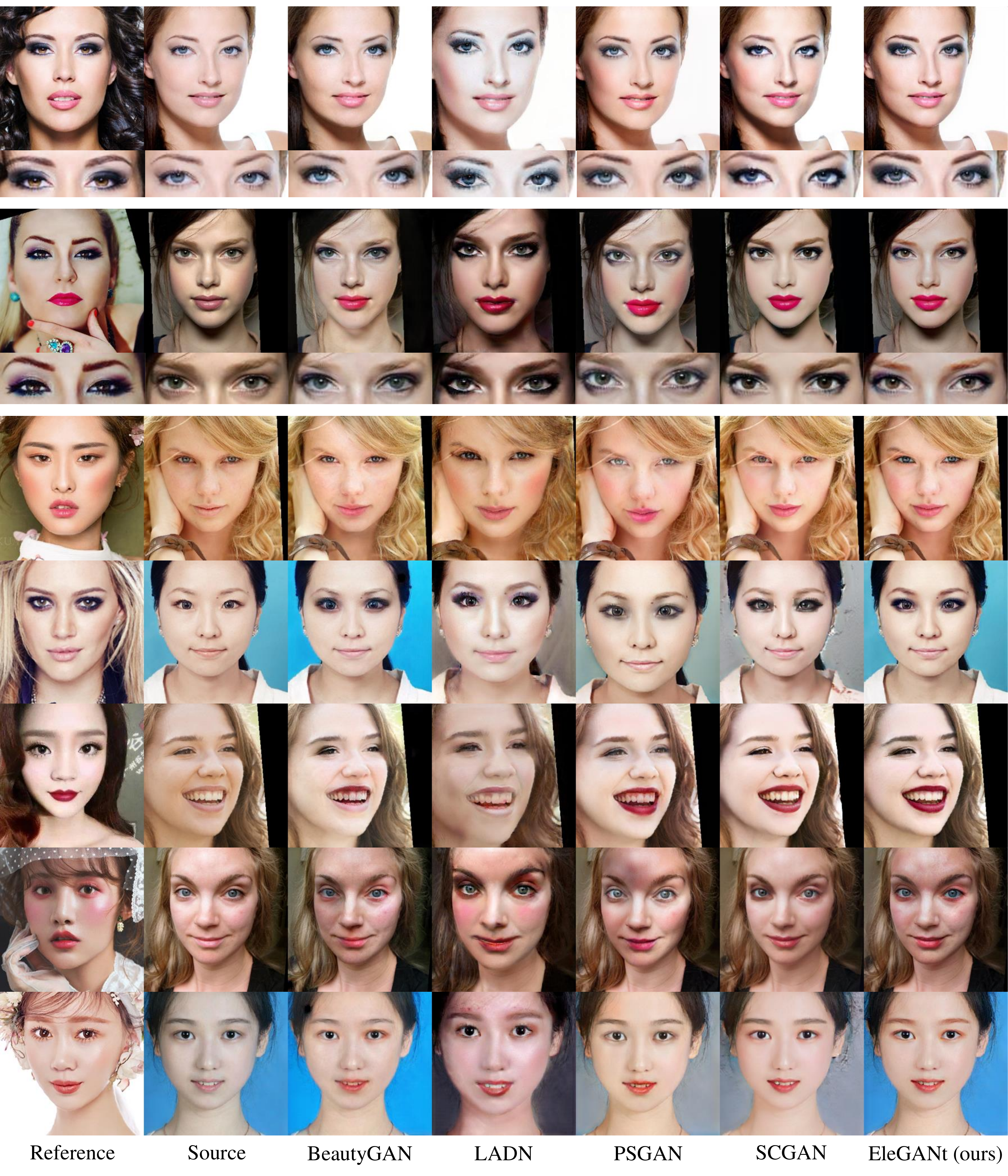}
    \caption{Qualitative comparisons with existing methods.}
    \label{fig:add_comparison}
\end{figure}
Fig. \ref{fig:add_comparison} presents additional qualitative results of BeautyGAN \cite{BeautyGAN}, LADN \cite{LADN}, PSGAN \cite{PSGAN}, SCGAN \cite{SCGAN} and our EleGANt. BeautyGAN generates visually acceptable results when the images have the same poses, but it falls short when a large spatial difference exists between the two faces. There are severe artifacts and blurs in the results of LADN, and the transferred colors are also incorrect. PSGAN cannot transfer detailed makeup attributes and suffers from unnatural shadows and illuminations. SCGAN fails to synthesize makeup details, and there are rectangular color blocks around the eyes in the generated images due to an improper decomposition that manually splits the face with rectangles. Our EleGANt surpasses all existing methods: it is robust to misaligned head poses and different illuminations, and it can preserve and precisely transfer makeup details, representatively, the shapes and colors of the eye shadows.

\end{document}